\documentclass[letter,12pt]{article}

\usepackage[letterpaper,margin=1in]{geometry}
\usepackage{authblk}
\usepackage{amsmath}
\usepackage{amsfonts}
\usepackage{rotating}
\usepackage{booktabs}
\usepackage[labelfont=bf]{caption}
\usepackage{setspace} 
\usepackage[english]{babel}
\usepackage{graphicx}
\usepackage[document]{ragged2e}
\usepackage{times}
\usepackage{algpseudocode}
\usepackage{algorithm}
\usepackage{amssymb}
\usepackage{bbm}

\usepackage{fancyhdr}
\usepackage{array}
\newcolumntype{P}[1]{>{\centering\arraybackslash}p{#1}}

\usepackage{xcolor}

\usepackage{amsmath}
\usepackage{hyperref}
\usepackage[style=nature,defernumbers=true,backend=biber]{biblatex}
\DeclareBibliographyCategory{inMain}
\addtocategory{inMain}{Griffiths2015}
\addtocategory{inMain}{gershman2018successor}
\addtocategory{inMain}{stachenfeld2017hippocampus}

\makeatletter
\let\blx@rerun@biber\relax
\makeatother

\renewcommand\Affilfont{\fontsize{11}{12}\itshape}
\makeatletter
\renewcommand\AB@affilsepx{; \protect\Affilfont}
\makeatother

\title{\Large \bf People construct simplified mental representations to plan}
\author[1,2,*]{Mark K. Ho}
\author[3,+]{David Abel} 
\author[4]{Carlos G. Correa}
\author[3]{Michael L. Littman}
\author[1,4]{Jonathan D. Cohen}
\author[1,2]{Thomas L. Griffiths}

\affil[1]{Princeton University, Department of Psychology, Princeton, NJ, USA}
\affil[2]{Princeton University, Department of Computer Science, Princeton, NJ, USA}
\affil[3]{Brown University, Department of Computer Science, Providence, RI, USA}
\affil[+]{Now at DeepMind, London, United Kingdom}
\affil[4]{Princeton University, Princeton Neuroscience Institute, Princeton, NJ, USA}
\affil[*]{Corresponding author: Mark K Ho, \texttt{mho@princeton.edu}}
\date{}

\begin{document}

\maketitle

\newrefsegment

\justifying
{\bf 
One of the most striking features of human cognition is the capacity to plan. Two aspects of human planning stand out: its efficiency and flexibility. Efficiency is especially impressive because plans must often be made in complex environments, and yet people successfully plan solutions to myriad everyday problems despite having limited cognitive resources~\supercite{lewis2014computational,Griffiths2015,Gershman2015computational}. Standard accounts in psychology, economics, and artificial intelligence have suggested human planning succeeds because people have a complete representation of a task and then use heuristics to plan future actions in that representation~\supercite{NewellSimon1972, russell2009artificial,keramati2016,Huys2012,Huys2015,callaway2018resource,sezener2019optimizing,pezzulo2019planning}. However, this approach generally assumes that task representations are \textit{fixed}. Here, we propose that task representations can be \textit{controlled} and that such control provides opportunities to quickly simplify problems and more easily reason about them. We propose a computational account of this simplification process and, in a series of pre-registered behavioral experiments, show that it is subject to online cognitive control~\supercite{miller2001integrative,shenhav2013expected,Shenhav2017} and that people optimally balance the complexity of a task representation and its utility for planning and acting. These results demonstrate how strategically perceiving and conceiving problems facilitates the effective use of limited cognitive resources.
}

\newpage

In the short story ``On Exactitude in Science,'' Jorge Luis Borges describes cartographers who seek to create the perfect map, one that includes every possible detail of the country it represents. However, this innocent premise leads to an absurd conclusion: The fully detailed map of the country must be the size of the country itself, which makes it impractical for anyone to use. Borges' allegory illustrates an important computational principle. Namely, useful representations do not simply mirror every aspect of the world, but rather pick out a manageable subset of details that are relevant to some purpose (Figure~1a). Here, we examine the consequences of this principle for how humans flexibly construct simplified task representations to plan.

Classic theories of problem solving distinguish between \textit{representing a task} and \textit{computing a plan}~\supercite{NewellSimon1972, norman1986attention,holland1989induction}. For instance, Newell and Simon~\supercite{newell1976computer} introduced \textit{heuristic search}, in which a decision-maker has a full representation of a task (e.g., a chess board, chess pieces, and the rules of chess), and then computes a plan by simulating and evaluating possible action sequences (e.g., sequences of chess moves) to find one that is likely to achieve a goal (e.g., checkmate the king). In artificial intelligence, the main approach to making heuristic search tractable involves limiting the computation of action sequences (e.g., only thinking a few moves into the future, or only examining moves that seem promising)~\supercite{russell2009artificial}. Similarly, psychological research on planning largely focuses on how limiting, prioritizing, pruning, or chunking action sequences can reduce computation~\supercite{daw2005uncertainty,glascher2010states,keramati2016,Huys2012,Huys2015,ramkumar2016chunking,callaway2018resource,sezener2019optimizing,pezzulo2019planning}.

However, people are not necessarily restricted to a single, full, or fixed representation for a task. This matters since simpler representations can make better use of limited cognitive resources when they are tailored to specific parts or versions of a task. For example, in chess, considering the interaction of a few pieces, or focusing on part of the board, is easier than reasoning about every piece and part of the board. Furthermore, it affords the opportunity to adapt the representation, tailoring it to the specific needs of the circumstance---a process that we refer to as \textit{controlling a task construal}. Although studies show that people can flexibly form representations to guide action (e.g., forming the \textit{ad hoc} category of ``things to buy for a party'' when organizing a social gathering~\supercite{barsalou1983ad}), a long-standing challenge for cognitive science and artificial intelligence is explaining, predicting, and deriving such representations from general computational principles~\supercite{Simon1975,brooks1991intelligence}. 

\begin{figure}[H]
    \centering
    \includegraphics[height=.65\textheight,trim=200 260 200 260,clip]{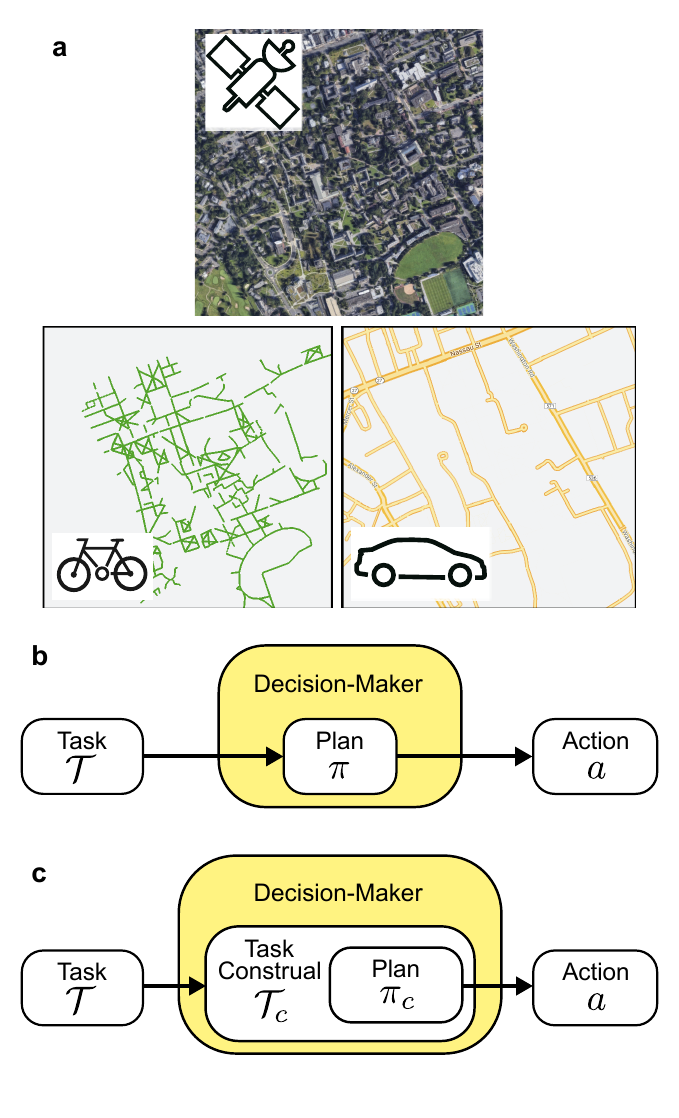}
    \caption*{
    \textbf{Figure 1. Construal and planning.} \textbf{a, } A satellite photo of Princeton, NJ (top) and maps of Princeton for bicycling versus automotive use cases (bottom). Like maps and unlike photographs, a decision-maker's \textit{construal} picks out a manageable subset of details from the world relevant to their current goals. Imagery \textcopyright 2022 Google, Map data \textcopyright 2022. \textbf{b, } Standard models assume that a decision-maker computes a plan, $\pi$, with respect to a fixed task representation, $\mathcal{T}$, and then uses it to guide their actions, $a$. \textbf{c, } According to our model of \textit{value-guided construal}, the decision-maker forms a simplified task construal, $\mathcal{T}_c$, that is used to compute a plan, $\pi_{c}$. This process can be understood as two nested optimizations: an ``outer loop'' of construal and an ``inner loop'' of planning. 
    }
\end{figure}

Our approach to studying how people control task construals starts with the premise that effective decision-making depends on making rational use of limited cognitive resources~\supercite{lewis2014computational,Griffiths2015,Gershman2015computational}. Specifically, we derive how an ideal, cognitively-limited decision-maker should form \textit{value-guided construals} that balance the complexity of a representation and its utility for planning and acting. We then show that pre-registered predictions of this account explain how people attend to task elements in several planning experiments (see Data Availability Statement). Our analysis and findings suggest that controlled, moment-to-moment task construals play a key role in efficient and flexible planning.

\section*{Task construals from first principles}

We build on models of sequential decision-making expressed as Markov Decision Processes~\supercite{puterman1994markov}. Formally, a task $\mathcal{T}$ consists of a state space, $\mathcal{S}$; an initial state, $s_0 \in \mathcal{S}$; an action space, $\mathcal{A}$; a transition function $P: \mathcal{S} \times \mathcal{A} \times \mathcal{S} \rightarrow [0, 1]$; and a utility function $U: \mathcal{S} \rightarrow \mathbb{R}$. In standard formulations of planning, the \textit{value} of a plan $\pi: \mathcal{S} \times \mathcal{A} \rightarrow [0, 1]$ from a state $s$ is determined by the expected, cumulative utility of using that plan~\supercite{bellman1957dynamic}:
$V_{\pi}(s) = U(s) + \sum_{a}\pi(a \mid s) \sum_{s'}P(s' \mid s, a) V_{\pi}(s')$. 
Standard planning algorithms~\supercite{russell2009artificial} (e.g., heuristic search methods) attempt to efficiently compute plans that optimize value by directly planning over a fixed task representation, $\mathcal{T}$, that is not subject to the decision-maker's control ({Figure~1b}). Our aim is to relax this constraint and consider the process of adaptively selecting simplified task representations for planning, which we call the \textit{construal} process ({Figure~1c}).

Intuitively, a construal ``picks out'' details in a task to consider. Here, we examine construals that pick out cause-effect relationships in a task. This focus is motivated by the intuition that a key source of task complexity is the interaction of different causes and their effects with one another. For instance, consider interacting with various objects in someone's living room. \textit{Walking towards the couch and hitting it} is a cause-effect relationship, while \textit{pulling on the coffee table and moving it} might be another such relationship. These individual effects can interact and may or may not be integrated into a single representation of moving around the living room. For example, imagine pulling on the coffee table and causing it to move, but in doing so, backing into the couch and hitting it. Whether or not a decision-maker anticipates and represents the interaction of multiple effects depends on what causes and effects are incorporated into their construal; this, in turn, can impact the outcome of behavior. 

Related work has studied how attention guides learning about how different \textit{state features} predict rewards~\supercite{leong2017dynamic}. By contrast, to model construals, we require a way to express how attention flexibly combines different \textit{causes and their effects} into an integrated model to use for planning. For this, we use a \textit{product of experts}~\supercite{hinton1999products}, a technique from the machine learning literature for combining distributions that is similar to factored approximations used in models of perception~\supercite{whiteley2012attention}. Specifically, we assume that the agent has $N$ primitive cause-effect relationships that each assign probabilities to state, action, and next-state transitions, $\phi_i: \mathcal{S} \times \mathcal{A} \times \mathcal{S} \rightarrow [0, 1]$, $i = 1, ..., N$. Each $\phi_i(s' \mid s, a)$ is a potential function representing, say, the local effect of colliding with the couch or pulling on the coffee table. Then a \textit{construal} is a subset of these primitive cause-effect relationships, $c \subseteq \{\phi_1, ..., \phi_N\}$, that produces a task construal, $\mathcal{T}_{c}$, with the following construed transition function:
\begin{equation}
P_c(s' \mid s, a) \propto \prod_{\phi_i \in c} \phi_i(s' \mid s, a).
\label{eq:construedtf}
\end{equation}
Here, we assume that task construals ($\mathcal{T}_c$) and the original task ($\mathcal{T}$) share the same state space, action space, and utility function. But, crucially, the construed transition function can be \textit{simpler} than that of the actual task.

What task construal should a decision-maker select? Ideally, it would be one that only includes those elements (cause-effect relationships) that lead to successful planning, excluding any others so as to make the planning problem as simple as possible. To make this intuition precise, it is essential to first distinguish between \textit{computing a plan} with a construal and \textit{using the plan} induced by a construal. In our example, suppose the decision-maker forms a construal of their living room that includes the effect of pulling on the coffee table but ignores the effect of colliding with the couch. They might then compute a plan in which they pull on the coffee table without any complications, but when they \textit{use} that plan in the actual living room, they inadvertently stumble over their couch. This particular construal is less than optimal.

Thus, we formalize the distinction between the \textit{computed plan} associated with a construal and its resulting \textit{behavioral utility}: If the decision-maker has a task construal $\mathcal{T}_c$, denote the plan that optimizes it as $\pi_{c}$. Then, the utility of the computed plan when starting at state $s_0$ is given by its performance when interacting with the actual transition dynamics, $P$:
\begin{equation}
U(\pi_c) = U(s_0) + \sum_{a} \pi_{c}(a \mid s_0)\sum_{s'}P(s' \mid s_0, a)V_{\pi_{c}}(s').
\label{eq:actualvalue}
\end{equation}
Put simply, the behavioral utility of a construal is determined by the consequences of using it to plan and act in the actual task.

Having established the relationship between a construal and its utility, we can define the \textit{value of representation} (VOR) associated with a construal. Our formulation resembles previous models of resource-rationality~\supercite{Griffiths2015} and the expected value of control~\supercite{shenhav2013expected} by discounting utilities with a \textit{cognitive cost}, $C$. This cost could be further enriched by specifying algorithm-specific costs~\supercite{Lieder2020} or hard constraints~\supercite{yoo2018strategic}. However, our aim is to understand value-guided construal with respect to the complexity of the construal itself and with minimal algorithmic assumptions. To this end, we use a cost that penalizes the number of effects considered: $C(c) = |c|$, where $|c|$ is the cardinality of $c$. Intuitively, this cost reflects the \textit{description length} of a program that expresses the construed transition function in terms of primitive effects~\supercite{grunwald2000model}. It also generalizes recent economic models of sparsity-based \textit{behavioral inattention}~\supercite{gabaix2014sparsity}. The value of representation for construal $c$ is then its behavioral utility minus its cognitive cost:
\begin{equation}
\text{VOR}(c) = U(\pi_c) - C(c).
\label{eq:voc}
\end{equation}

In short, we introduce the notion of a task construal (Equation~\ref{eq:construedtf}) that relaxes the assumption of planning over a fixed task representation. We then define an optimality criterion for a construal based on its complexity and its utility for planning and acting (Equations~\ref{eq:actualvalue}-\ref{eq:voc}). This optimality criterion provides a normative standard we can use to ask whether people form optimal value-guided construals~\supercite{marr1982vision,Anderson1990}. We note that the question of precisely how people identify or learn optimal construals is beyond the scope of our current aims. Rather, here our goal is to simply determine whether their planning is consistent with optimal construal. If so, then understanding how people achieve (or approximate) this ability will be a key direction for future research (see Supplementary Discussion of Construal Optimization Algorithms).

\section*{A paradigm for examining construals}
Do people form construals that optimally balance complexity and utility? To answer this question, we designed a paradigm analogous to the example in {Figure~1a}, in which participants were shown a two-dimensional map of a maze and had to move a blue dot to reach a goal location. On each trial, participants were shown a new maze composed of a starting location, a goal location, center black walls in the shape of a $+$, and an arrangement of blue obstacles. The goal, starting state, and the blue obstacles (but not the center black walls) changed on every trial, which required participants to examine the layout of the maze and plan an efficient route to the goal (Figure~2a). In our framework, each obstacle corresponds to a cause-effect relationship, $\phi_i$---i.e., attempting to move into the space occupied by the obstacle and then being blocked. This is analogous to the effect of being blocked by a piece of furniture in our earlier example.

Two key features make our maze-navigation paradigm useful for isolating and studying the construal process. First, the mazes are \textit{fully observable}: Complete information about the task is immediately accessible from the visual stimulus. Second, each instance of a maze emerges from a particular \textit{composition} of individual elements (e.g., the obstacles). This means that while all the components of a particular maze are immediately accessible, participants need to choose which ones to \textit{integrate} into an effective representation for planning (i.e., select a construal). Fully observable but compositionally-structured problems occur routinely in everyday life---e.g., using a map to navigate through exhibits in a museum---as well as in popular games---e.g., in chess, figuring out how to move one's knight across a board occupied by an opponent's pieces. By providing people with immediate access to all the components of a task while planning, we can examine which ones they \textit{attend to} versus \textit{ignore} and whether these patterns of awareness reflect a process of value-guided construal (Methods, Model Implementations, Value-guided Construal Implementation; Code Availability Statement). Furthermore, this general paradigm can be used in concert with several different experimental measures to assess attention (Extended Data Figures 1-3; Supplementary Experimental Materials; Data Availability Statement).

\begin{figure}[H]
    \centering
    \includegraphics[width=\textwidth,trim=80 280 80 280,clip]{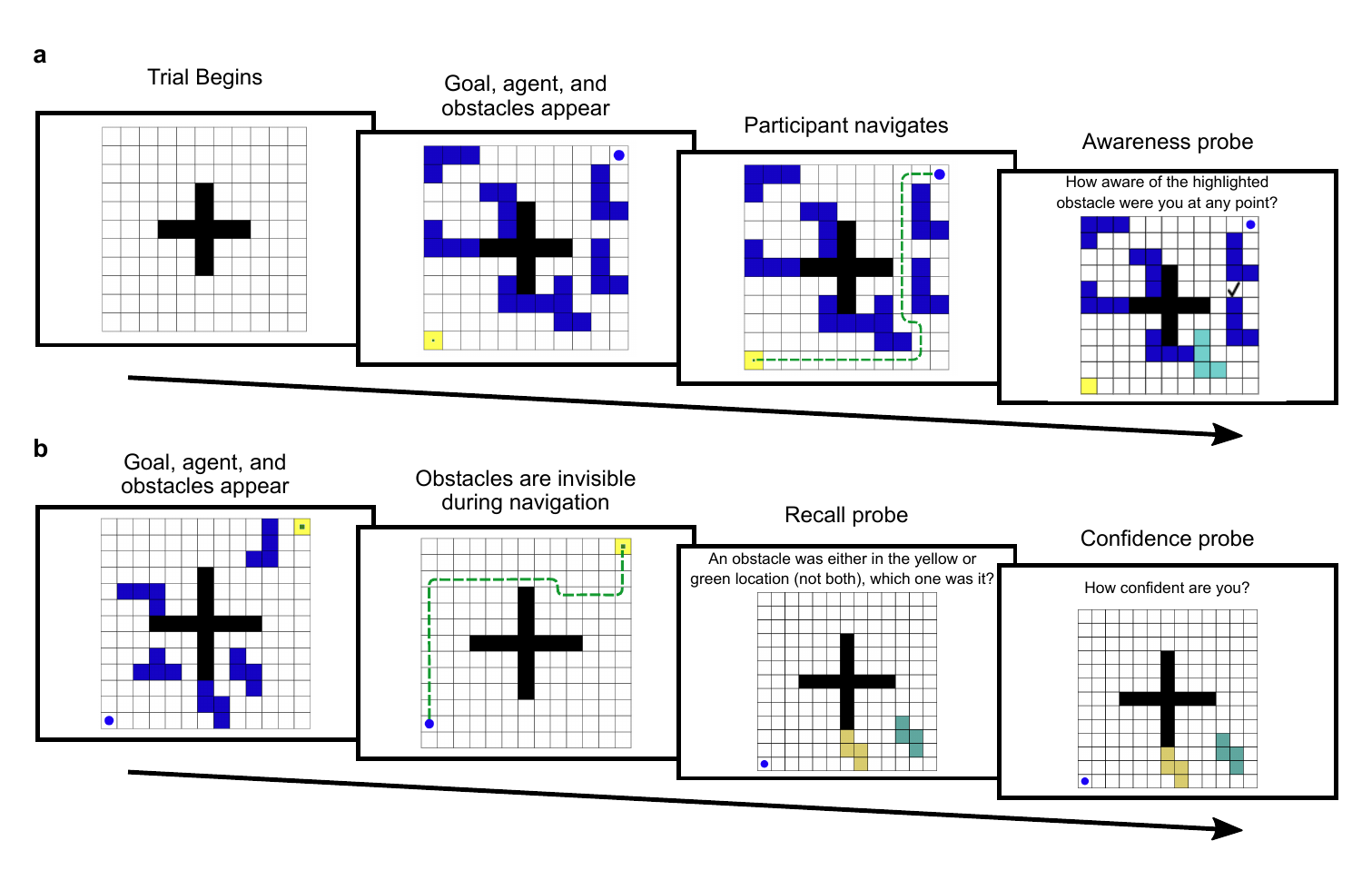}
    \caption*{
    \textbf{Figure 2. Maze-navigation paradigm and design of memory probes,} Value-guided construal predicts how people will form representations that are simple but useful for planning and acting. These predictions were tested in a new paradigm in which participants controlled a blue circle and navigated mazes composed of center black walls in the shape of a cross, blue tetronimo-shaped obstacles, and a yellow goal state with a shrinking green square. We assume that attention to obstacles as a result of construal is reflected in memory of obstacles and used two types of probes to assess memory. \textbf{a, } In our initial experiment, participants were shown the maze and navigated to the goal (dashed line indicates an example path). After navigating, participants were given \textit{awareness probes} in which they were asked to report their awareness of each obstacle on an 8-point scale (for analyses, responses were scaled to range from 0 to 1). \textbf{b, } In a subsequent experiment, obstacles were only visible prior to moving in order to encourage planning up-front, and participants were given \textit{recall probes} in which they were shown a pair of obstacles in green and yellow, only one of which had been present in the maze they had just completed. They were then asked which one had been in the maze as well as their confidence.
    }
\end{figure}

\section*{Traces of construals in people's memory}
We assume that the obstacles included in a construal will be associated with greater awareness and thereby memory; accordingly, we began by probing memory for obstacles after participants completed each maze to test whether they formed value-guided construals of the mazes. In our initial experiment, participants received \textit{awareness probes} in which, following navigation, they were shown a picture of the maze they had just completed with one of the obstacles highlighted. Then, they were asked, ``How aware of the highlighted obstacle were you at any point?'' and responded on an 8-point scale that was later scaled to range from 0 to 1 for analyses (Figure~2a). If participants formed representations of the mazes that balance utility and complexity, their responses should be positively predicted by value-guided construal. This is precisely what we found: Value-guided construal predicted awareness judgments (likelihood ratio test comparing hierarchical linear models with and without z-score normalized value-guided construal probabilities: $\chi^2(1) = 2297.21, p  < 1.0 \times 10^{-16}$; $\beta = 0.133$, S.E. $= 0.003$\unskip; Methods, Experiment Analyses; Figure~3). Furthermore, we also observed the same results when participants could not see the obstacles while moving and so needed to plan their route entirely up front ($\chi^2(1) = 726.95, p  < 1.0 \times 10^{-16}$; $\beta = 0.115$, S.E. $= 0.004$\unskip). This was also the case when we probed awareness judgments \textit{immediately after planning but before execution} ($\chi^2(1) = 679.20, p  < 1.0 \times 10^{-16}$; $\beta = 0.106$, S.E. $= 0.004$\unskip; Methods, Experimental Design, Up-front Planning Experiment; Supplementary Memory Experiment Analyses).

\begin{figure}[H]
    \centering
    \includegraphics[width=\textwidth,trim=40 400 40 160,clip]{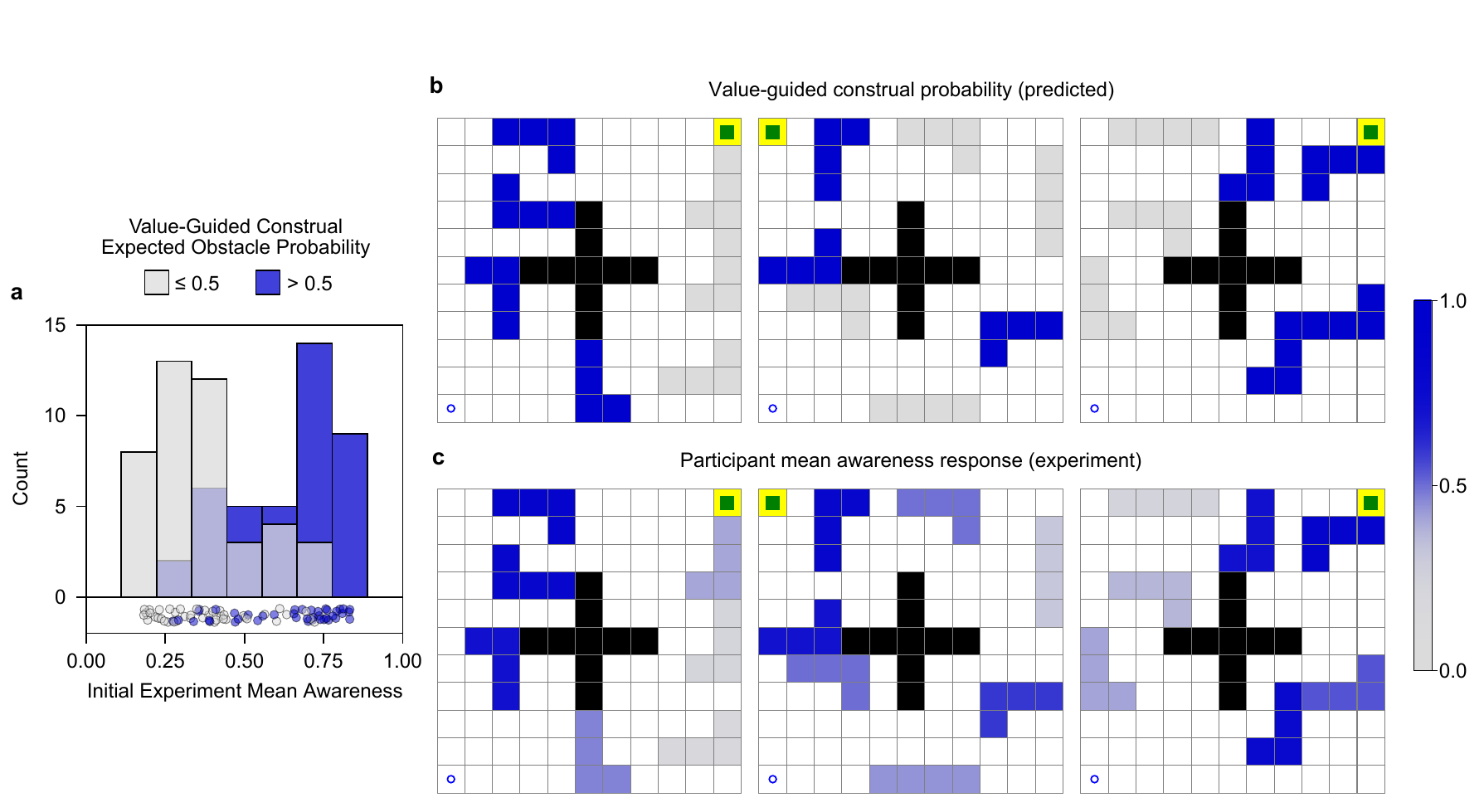}
    \caption*{
    \textbf{Figure 3. Initial experiment results,} In our initial planning experiment (out of four), each person ($n=161$ independent participants) navigated twelve 2D mazes, each of which had seven blue tetronimo-shaped obstacles. To assess whether attention to obstacles reflects a process of value-guided construal, participants were given an awareness probe (see Figure 2a) for each obstacle in each maze. \textbf{a, } For our first analysis, we split the set of 84 obstacles across mazes based on whether value-guided construal assigned a probability less than or equal to $0.5$ or greater than $0.5$. Here, we plot two histograms of participants' mean awareness responses corresponding to the two sets of obstacles ($\leq\!0.5$ in grey, $>\!0.5$ in blue; individual by-obstacle mean awareness underlying the histograms are represented underneath). We then similarly split the obstacles based on whether mean awareness responses were less than or equal to $0.5$ or greater than $0.5$ and, using a chi-squared test for independence, found that this split was predicted by value-guided construal (\input{inputs/exp1_chi2_res_svgc}\unskip). \textbf{b, } Value-guided construal predictions for three of the twelve mazes used in the experiment (blue circle indicates the starting location, green and yellow square indicates the goal; obstacle colors represent model probabilities according to the colorbar). \textbf{c, } Participant mean awareness judgments for the same three mazes (obstacle colors represent mean judgments according to the colorbar). Responses in this initial experiment generally reflect value-guided construal of mazes. Participants were recruited through the Prolific online experiment platform.
    }
\end{figure}

While the awareness probes provide useful insight into people's task construals, it is a step removed from their memory (which is already a step removed from the construal process itself) since it requires participants to reflect on their earlier awareness during planning. To address this limitation, we developed a second set of \textit{critical mazes} with two properties. First, the mazes were designed to test the distinctive predictions of value-guided construal (e.g., Figure~4a). Second, these new mazes allowed us to use a more stringent measure of memory for task elements. Specifically, we used \textit{obstacle recall probes}, in which, following navigation, participants were shown a grid with the black center walls, a green obstacle, a yellow obstacle, and no other obstacles. Either the green or yellow obstacle had actually been present in the maze, whereas the other obstacle did not overlap with any of those that had been present. Participants were then asked, ``An obstacle was either in the yellow or green location (not both), which one was it?'' and could select either option, followed by a confidence judgment on an 8-point scale (Figure~2b; Extended Data Figure 4a). The recall probes thus provided two measures, accuracy and confidence, and using hierarchical generalized linear models (HGLMs) we found that value-guided construal predicted both types of responses (likelihood ratio tests comparing models on accuracy: $\chi^2(1) = 249.34, p  < 1.0 \times 10^{-16}$; $\beta = 0.648$, S.E. $= 0.042$\unskip; and confidence: $\chi^2(1) = 432.76, p  < 1.0 \times 10^{-16}$; $\beta = 0.104$, S.E. $= 0.005$\unskip. Methods, Experiment Analyses). Additionally, when we gave a separate group of participants the awareness probes on these mazes, value-guided construal was again predictive (Awareness: $\chi^2(1) = 837.47, p  < 1.0 \times 10^{-16}$; $\beta = 0.175$, S.E. $= 0.006$\unskip). Thus, using three different measures of memory (recall accuracy, recall confidence, and awareness judgments), we found further evidence that when planning, people form task representations that optimally balance complexity and utility.

\begin{figure}[H]
    \centering
    \includegraphics[width=.95\textwidth,trim=0 0 0 0,clip]{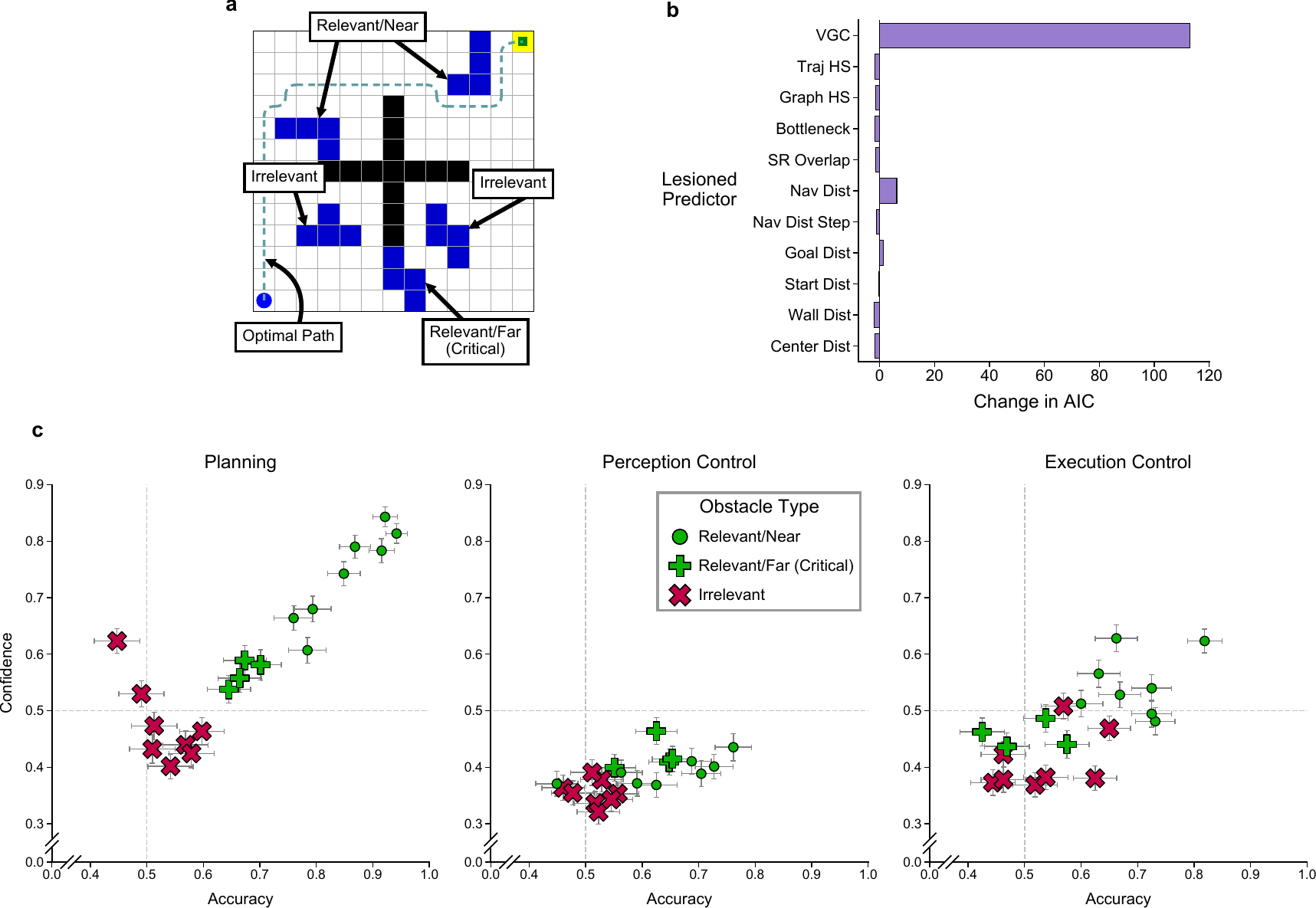}
    \caption*{
    \textbf{Figure 4. Critical mazes recall experiment, model comparisons, and control studies.} \textbf{a,} The critical mazes recall experiment ($n=78$ independent participants; one version of one of the four planning experiments) used \textit{critical mazes} that included \textit{critical obstacles} that were highly relevant to planning but far from an optimal path (dashed line). Value-guided construal predicts critical obstacles will be included in a construal while \textit{irrelevant obstacles} will not, independent of distance to the optimal path. \textbf{b,} We fit a global model to recall responses that included the fixed parameter value-guided construal modification model (VGC) along with ten alternative predictors based on heuristic search models, successor representation-based predictors, and low-level perceptual cues (see Methods, Experiment Analyses). Then, each predictor was removed from this global model, and we calculated the resulting change in fit (in AIC). Removing value-guided construal led to the largest degradation of fit (greatest increase in AIC), underscoring its unique explanatory value. \textbf{c,} In a pair of non-planning control experiments, new participants either viewed patterns that looked exactly like the mazes (perceptual control; $n=88$ independent participants) or followed ``breadcrumbs'' through the maze along a path taken by a participant from the original experiment (execution control; $n=80$ independent participants). They then answered the exact same recall questions. Value-guided construal remains a significant factor when explaining recall in the original critical mazes experiment (planning) while including mean recall from the perceptual and execution controls as covariates (likelihood ratio test for accuracy: \protect\input{inputs/exp3_acc_control_vgc_test_svgc}\unskip; confidence: \protect\input{inputs/exp3_conf_control_vgc_test_svgc}\unskip; $p$-values are unmodified). This confirms that responses consistent with value-guided construal are not a simple function of perception and execution. Participants were recruited through the Prolific online experiment platform. Plotted are the mean values for each obstacle, with relevant/near, relevant/far (critical), and irrelevant obstacle types distinguished. Error bars are standard errors about the mean.
    }
\end{figure}

\section*{Controlling for perception and execution}
The memory studies provide preliminary confirmation of our hypothesis, but they have several limitations. One is that, although participants were engaged in \textit{planning}, they were also necessarily engaged in other forms of cognitive processing, and these unrelated processes may have influenced memory of the obstacles. In particular, participants' \textit{perception} of a maze or their \textit{execution} of a particular plan through a maze may have influenced their responses to the memory probes. This potentially confounds the interpretation of our results, since a key part of our hypothesis is that task construals arise from \textit{planning}, rather than simply perceiving or executing.

Thus, to test that responses to the memory probes cannot be fully explained by perception and/or execution, we administered two sets of \textit{yoked controls} that did not require planning (Methods, Experimental Design, Control Experiments). In the \textit{perceptual controls}, new participants were shown patterns that looked exactly like the mazes, but they performed an unrelated, non-planning task. Each pattern was presented to a new participant for the same amount of time that a participant in the original experiments had examined the corresponding maze before moving---i.e., the amount of time the original participant spent examining the maze to plan. The new participant then responded to the same probes, in the same order, as the original participant. For the \textit{execution controls}, we recruited another group of participants and gave them instructions similar to those in the planning experiments. However, unlike the original experiments, the task did not require planning. Rather, these mazes included ``breadcrumbs'' that needed to be collected and that appeared every two steps. Breadcrumbs appeared along the exact path taken by one of the original participants, meaning that the new participant executed the same actions but \textit{without having planned}. After completing each maze, the participant then received the same probes, in the same order, as the original participant. 

We assessed whether responses in the planning experiments can be explained by a simple combination of perception and/or execution by testing whether value-guided construal remained a significant factor after accounting for control responses. Specifically, we used the mean by-obstacle responses from the perceptual and execution controls as predictors in HGLMs fit to the corresponding planning responses. We then tested whether adding value-guided construal as a predictor improved fits. For the awareness, accuracy, and confidence responses in the recall experiment, we found that including value-guided construal significantly improved fits (likelihood ratio tests comparing models on accuracy: $\chi^2(1) = 106.36, p = 6.2 \times 10^{-25}$\unskip; confidence: $\chi^2(1) = 18.56, p = 1.6 \times 10^{-5}$\unskip; and awareness: $\chi^2(1) = 55.34, p = 1.0 \times 10^{-13}$\unskip) and that value-guided construal predictions were positively associated with responses (coefficients for accuracy: $\beta = 0.58, \text{S.E.} = 0.058$\unskip; confidence: $\beta = 0.039, \text{S.E.} = 0.009$\unskip; and awareness: $\beta = 0.054, \text{S.E.} = 0.007$\unskip). Thus, responses following \textit{planning} are not reducible to a simple combination of \textit{perception} and \textit{execution}, and they can be further explained by the formation of value-guided construals (Figure~4c; Supplementary Control Experiment Analyses).

\section*{Externalizing the planning process}
Another limitation of the previous planning experiments is that they assess construal \textit{after} planning is complete (i.e., by probing memory). To obtain a measure of the planning process \textit{as it unfolds}, we developed a novel \textit{process-tracing paradigm}. In this version of the task, participants never saw all of the obstacles at once. Instead, at the beginning of the trial, after being shown the start and goal locations, they could use their mouse to reveal individual obstacles by hovering over them (Methods, Experimental Design, Process-tracing Experiments; Extended Data Figure 4b). This led participants to externalize the planning process, and so their behavior on this task provides insight into how planning computations unfolded internally. We tested whether value-guided construal accounted for behavior by analyzing two measures: whether an obstacle was hovered over and, if it was hovered over, the duration of hovering. Value-guided construal was a significant predictor for both these measures on both the initial mazes (likelihood ratio tests comparing HGLMs for hovering: $\chi^2(1) = 1221.76, p  < 1.0 \times 10^{-16}$; $\beta = 0.704$, S.E. $= 0.021$\unskip; and hover duration [log milliseconds]: $\chi^2(1) = 169.90, p  < 1.0 \times 10^{-16}$; $\beta = 0.161$, S.E. $= 0.012$\unskip) and on the critical mazes (hovering: $\chi^2(1) = 1361.92, p  < 1.0 \times 10^{-16}$; $\beta = 0.802$, S.E. $= 0.023$\unskip; hover duration [log milliseconds]: $\chi^2(1) = 540.63, p  < 1.0 \times 10^{-16}$; $\beta = 0.369$, S.E. $= 0.016$\unskip). These results thus complement our original memory-based measurements of people's task representations and strengthen the interpretation of them in terms of value-guided construal during planning.

\section*{Value-guided construal modification}
Thus far, our account of value-guided construal has assumed that an obstacle is either always or never included in a construal. This simplification is useful since it enables us to derive clear qualitative predictions based on \textit{whether} a plan is influenced by an obstacle, but it overlooks graded factors such as \textit{how much} of a plan is influenced by an obstacle. For example, an obstacle may only be relevant for planning a few movements around a participant's initial location in a maze and, as a result, could receive less total attention than one that is relevant for deciding how to act across a larger area of the maze. To characterize these more fine-grained attentional processes, we first generalized the original construal selection problem to a one in which the decision-maker revisits and potentially modifies their construal during planning. Then, we derived obstacle awareness predictions based on a theoretically optimal \textit{construal modification policy} that balances complexity and utility (Methods, Model Implementation, Value-Guided Construal).

To assess value-guided construal modification, we re-analyzed our data using three versions of the model with increasing ability to capture variability in responses. First, we used an idealized \textit{fixed parameter} model to derive a single set of obstacle attention predictions and confirmed that they also predict participant responses on the planning tasks (Supplementary Construal Modification Analyses). Second, for each planning measure and experiment, we calculated \textit{fitted parameter} models in which noise parameters for the computed plan and construal modification policy were fit (Methods, Model Implementation, Value-Guided Construal). Scatter plots comparing mean by-obstacle responses and model outputs for parameters with the highest $R^2$ are shown in Figure 5. Finally, we fit a set of models that allowed for biases in computed plans (e.g., a bias to stay along the edge of a maze or an explicit penalty for bumping into walls) and found that this additional expressiveness led to obstacle attention predictions with an improved correspondence to participant responses (Supplementary Construal Modification Analyses). Together, these analyses provide additional insight into the fine-grained dynamic structure of value-guided construal modification.

\begin{figure}[H]
    \centering
    \includegraphics[width=\textwidth,trim=0 0 0 0,clip]{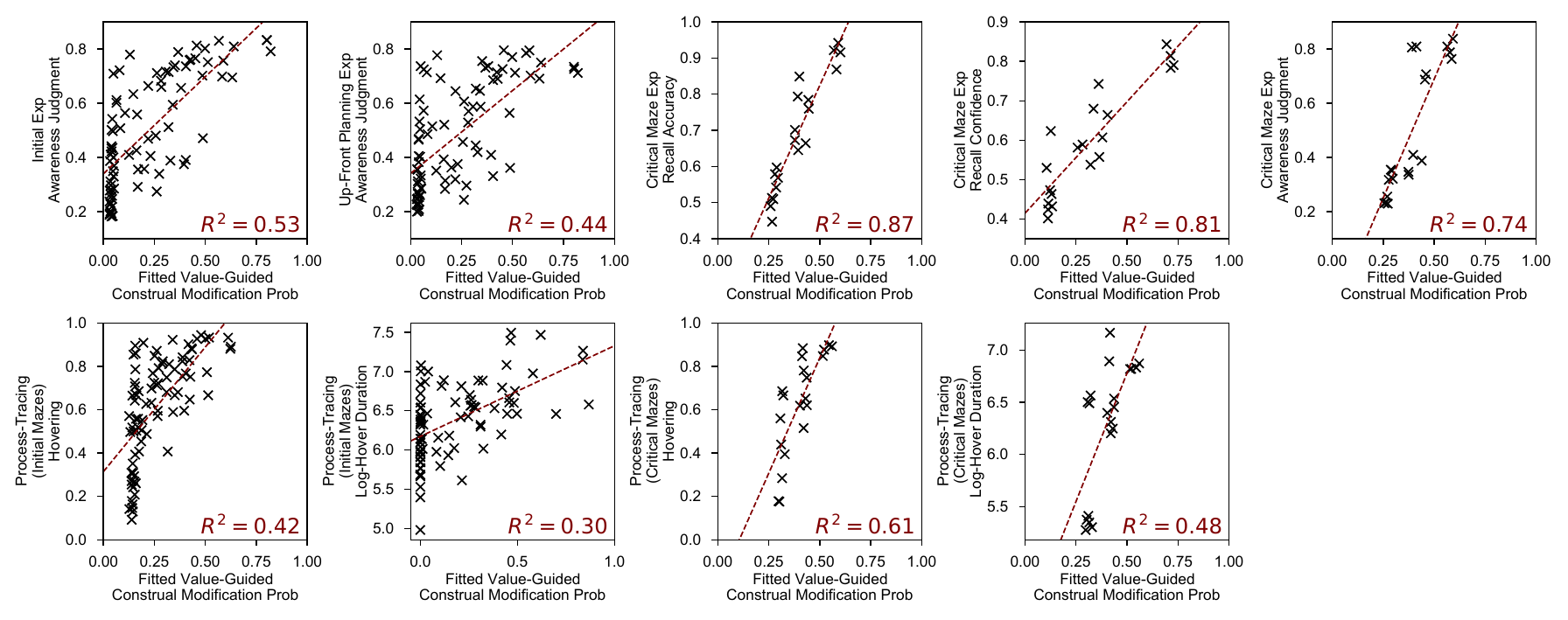}
    \caption*{
    \textbf{Figure 5. Fitted value-guided construal modification.} Our initial model of value-guided construal focuses on \textit{whether} an obstacle should or should not be included in a construal. We developed a generalization that additionally accounts for \textit{how much} an obstacle influences a plan if a decision-maker is optimally modifying their construal during planning (Methods, Model Implementations, Value-Guided Construal). We used an $\varepsilon$-softmax noise model~\cite{nassar2016taming} for computed action plans and construal modification policies and, for each experiment and measure, searched for parameters that maximize the $R^2$ between model predictions and mean by-obstacle responses. Shown here are plots comparing scores that the fitted construal modification model assigns to each obstacle with participants' mean by-obstacle responses for the nine measures.
    }
\end{figure}

\section*{Accounting for alternative mechanisms}
While the analyses so far confirm the predictive power of value-guided construal, it is also important to consider alternative planning processes. For instance, differential awareness could have been a \textit{passive side-effect of planning computations}, rather than an \textit{active facilitator of planning computations} as posited by value-guided construal. In particular, participants could have been planning by performing heuristic search over action sequences without actively construing the task, which would have led to differential awareness of obstacles as a byproduct of planning. Differential awareness could also have arisen from alternative representational processes, such as those based on the successor representation~\supercite{gershman2018successor} or related subgoaling mechanisms~\supercite{stachenfeld2017hippocampus}. Similarly, perceptual factors, such as the distance to the start, goal, walls, center, optimal path, or path taken, could have influenced responses. 

Based on these considerations, we identified ten alternative predictors (Methods, Model Implementations; Extended Data Figures 5, 6, and 7; Code Availability Statement). All ten predictors plus the fixed value-guided construal modification predictions were included in global models that were fit to each of the nine planning experiment measures, and, in all cases, value-guided construal was a significant predictor (Extended Data Table 1; see Supplementary Alternative Mechanisms Analyses for the same analyses with the single-construal model).

Furthermore, to assess the relative importance of each predictor, we calculated the change in fit (in terms of AIC) that resulted from removing each predictor from a global model (Methods, Experiment Analyses). Across all planning experiment measures, removing value-guided construal led to the first or second largest reduction in fit (Figure~4b; Extended Data Table 1). These ``knock-out'' analyses demonstrate the explanatory \textit{necessity} of value-guided construal. To assess explanatory \textit{sufficiency}, we fit a new set of single-predictor and two-predictor models using all predictors and then calculated their $\Delta$AICs (Methods, Experiment Analyses; Extended Data Figure 8). For all nine experimental measures, value-guided construal was one of the top two single-predictor models and was one of the two factors included in the best two-predictor model. Together, these analyses confirm the explanatory necessity and sufficiency of value-guided construal.

\section*{Discussion}
We tested the idea that when people plan, they do so by constructing a simplified mental representation of a problem that is sufficient to solve it---a process that we refer to as value-guided construal. We began by formally articulating how an ideal, cognitively-limited decision-maker should construe a task so as to balance complexity and utility. Then, we showed that pre-registered predictions of this model explain people's awareness, ability to recall problem elements (obstacles in a maze), confidence in recall ability, and behavior in a process-tracing paradigm, even after controlling for the baseline influence of perception and execution as well as ten alternative mechanisms. These findings support the hypothesis that people make use of a controlled process of value-guided construal, and that it can help explain the efficiency of human planning. More generally, our account provides a framework for further investigating the cognitive mechanisms involved in construal. For instance, how are construal strategies acquired? How is construal selection shaped by computation costs, time, or constraints? From a broader perspective, our analysis suggests a deep connection between the control of construals and the acquisition of structured representations like objects and their parts that can be cognitively manipulated~\supercite{tversky1984objects,tenenbaum2011grow}, which can inform the development of intelligent machines. Future investigation into these and other mechanisms that interface with the control of representations will be crucial for developing a comprehensive theory of flexible and efficient intelligence.

\printbibliography[title={Main References},segment=\therefsegment]

\newpage

\newrefsegment
\section*{Methods}

\subsection*{Model Implementations}

\subsubsection*{Value-guided Construal} 
Our model assumes the decision-maker has a set of cause-effect relationships that can be combined into a task construal that is then used for planning. To derive empirical predictions for the maze tasks, we assume a set of primitive cause-effect relationships, each of which is analogous to the example of interacting with furniture in a living room (see main text). For each maze, we modeled the following: The default effect of movement (i.e., pressing an arrow key causes the circle to move in that direction with probability $1 - \varepsilon$ and stay in place with probability $\varepsilon$, $\varepsilon = 10^{-5}$), $\phi_{\text{Move}}$; the effect of being blocked by the center, plus-shaped ($+$) walls (i.e., the wall causes the circle to \textit{not} move when the arrow key is pressed), $\phi_{\text{Walls}}$; and effects of being blocked by each of the $N$ obstacles, $\phi_{\text{Obstacle}_i}, i = 1, ..., N$. Since every maze includes the same movements and walls, the model only selected which obstacle effects to include. The utility function for all mazes was given by a step cost of $-1$ until the goal state was reached.

Value-guided construal posits a bilevel optimization procedure involving an ``outer loop'' of construal and an ``inner loop'' of planning. Here, we exhaustively calculate potential solutions to this nested optimization problem by enumerating and planning with all possible construals (i.e., subsets of obstacle effects). We exactly solved the inner loop of planning for each construal using dynamic programming~\supercite{sutton2018} and then evaluated the optimal stochastic computed plan under the actual task dynamics (i.e., Equation~\ref{eq:actualvalue}). For planning and evaluation, transition probabilities were multiplied by a discount rate of $.99$ was used to ensure values were finite. The general procedure for calculating the value of construals is outlined in the algorithm in Extended Data Table 2. To be clear, our current research strategy is to derive theoretically optimal predictions for the inner loop of planning and outer loop of construal in the spirit of resource-rational analysis~\supercite{Griffiths2015}. Thus, this specific procedure should not be interpreted as a process model of human construal. In the Supplemental Discussion of Algorithms for Construal Optimization, we discuss the feasibility of optimizing construals and how an important direction for future research will involve investigating tractable algorithms for finding good construals.

Given a value of representation function, $\text{VOR}$, that assigns a value to each construal, we model participants as selecting a construal according to a softmax decision-rule:
\begin{equation}
    P(c) \propto \exp \left\{ \alpha^{-1} \text{VOR}(c) \right\},
\label{eq:construalsoftmax}
\end{equation}
where $\alpha > 0$ is a temperature parameter (for our pre-registered predictions $\alpha = 0.1$). We then calculated a marginalized probability for each obstacle being included in the construal, from the initial state, $s_0$, corresponding to the expected awareness of that obstacle:
\begin{equation}
    P(\text{Obstacle}_i) = \sum_{c} \mathbbm{1}[\phi_{\text{Obstacle}_i} \in c] P(c),
\label{eq:obstacle-marginalize}
\end{equation}
where, for a statement $X$, $\mathbbm{1}[X]$ evaluates to $1$ if $X$ is true and $0$ if $X$ is false. We implemented this model in Python 3.7 using the \texttt{msdm} library (see Code Availability Statement).

The basic value-guided construal model makes the simplifying assumption that the decision-maker plans with a single static construal. We can extend this idea to consider a decision-maker who revisits and potentially modifies their construal at each stage of planning. In particular, we can conceptualize this process in terms of a sequential decision-making problem induced by the interaction between task dynamics (e.g., a maze) and the internal state of an agent (e.g., a construal)~\cite{parr1997reinforcement}. The solution to this problem is then a sequence of modified construals associated with planning over different parts of the task (e.g., planning movements for different areas of the maze). 

Formally, we denote the set of possible construals as $\mathcal{C} = \mathcal{P}(\{\phi_1, ..., \phi_N\})$, the powerset of cause-effect relationships, and define a \textit{construal modification Markov Decision Process}, which has a state space corresponding to the Cartesian product of task states and construals, $(s, c) \in \mathcal{S} \times \mathcal{C}$, and an action space corresponding to possible next construals, $c' \in \mathcal{C}$. Having chosen a new construal $c'$, the probability of transitioning from task state $s$ to $s'$ comes from first calculating a joint distribution using the actual transition function $P(s' \mid s, a)$ and plan $\pi_{c'}(a \mid s)$ and then marginalizing over task actions $a$:
\begin{equation}
    P(s' \mid s, c') = \sum_{a}  \pi_{c'}(a \mid s)P(s' \mid s, a).
\end{equation}
In this construal modification setting, the analogue to the value of representation (VOR; Equation~\ref{eq:voc}) is the \textit{optimal construal modification value function}, defined over all $s, c$:
\begin{equation}
    V(s, c) = U(s) + \max_{c'} \left[ \sum_{s'}P(s' \mid s, c')V(s', c') - C(c', c) \right],
\end{equation}
where $C(c', c) = |c' - c|$ is the number of \textit{additional}\footnote{For sets $A$ and $B$, the \textit{set difference} $A - B = \{a : a \in A \text{ and } a \notin B\}$.} cause-effect relationships in the new construal $c'$ compared to $c$. Importantly, this cost on modifying the construal encourages consistency---i.e., without $C(c', c)$, a decision-maker would have no disincentive to completely change their construal for each state. Note that in the special case where $c = \varnothing$, we recover the original static construal cost for a single step. Finally, using the construal modification value function, we define a softmax policy over the task/construal state space, $\pi(c' \mid s, c) \propto \exp \left\{\alpha_c^{-1}\left[\sum_{s'}P(s' \mid s, c')V(s', c') - C(c', c)\right]\right\}$. For the \textit{fixed parameter} model we set $\alpha_c = 0.1$ (as with the single-construal model).

The construal modification formulation allows us to consider not just \textit{whether} an obstacle appears in a construal, but also \textit{how long} it appears in a construal. In particular, we would like to compute a quantity that is analogous to Equation~\ref{eq:obstacle-marginalize} that assigns model values for each obstacle. To do this, we use the \textit{normalized task/construal state occupancy} induced by a construal policy $\pi$ from the initial task/construal state, $\rho_\pi(s, c \mid s_0, c_0) \propto M_\pi(s_0, c_0, s, c)$, where $c_0 = \varnothing$ and $M_\pi$ is the successor representation under $\pi$ (for a self-contained review of $M_\pi$, see the section on Successor Representation-based Predictors below). Given a policy $\pi$ and starting task state $s_0$, for each obstacle, we calculate the probability of having a construal that includes that obstacle:
\begin{equation}
    P(\text{Obstacle}_i) = \sum_{s, c}\mathbbm{1}[\phi_{\text{Obstacle}_i}\in c]\rho_\pi(s, c \mid s_0, c_0).
\end{equation}

To calculate the optimal construal modification value function, $V(s, c)$, for each maze, we constructed construal modification Markov Decision Processes in Python (3.7) using \texttt{scipy} (1.5.2) sparse matrices~\cite{scipy2020}. We then exactly solved for $V(s, c)$ using a custom implementation of policy iteration~\cite{howard1960dynamic} designed to take advantage of the sparse matrix data structure (see Code Availability Statement). For the \textit{fitted parameter} models, we used separate $\varepsilon$-softmax noise models~\cite{nassar2016taming} for the computed plans, $\pi_c(a \mid s)$, and construal modification policy, $\pi(c' \mid s, c)$, and performed a grid search over the four parameters for each of the nine planning measures ($\alpha_a^{-1} \in \{1, 3, 5, 7\}; \varepsilon_a \in \{0.0, 0.1, 0.2\}; \alpha_c^{-1} \in \{1, 3, 5, 7, 9\}; \varepsilon_c \in \{0, 0.05, 0.1, 0.2, 0.3\}$). Additionally, for parameter fitting, we limited the construals $c' \in \mathcal{C}$ to be of size three. This improves the speed of parameter evaluation and yields results comparable to the \textit{fixed parameter} model, which uses the full construal set. Finally, to obtain obstacle value-guided construal probabilities we simulate 1000 rollouts of the construal modification policy to estimate $\rho_\pi(\cdot \mid s_0, c_0)$. As with the initial model, we emphasize that these procedures are not intended as an algorithmic account of construal modification, but rather allow us to derive theoretically optimal predictions of the fine-grained dynamics of value-guided construals during planning.

\subsubsection*{Heuristic Search Over Action Sequences} 
Value-guided construal posits that people control their task representations to \textit{actively facilitate planning}, which, in the maze navigation paradigm, leads to differential attention to obstacles. However, differential attention could also occur as a \textit{passive side-effect of planning}, even in the absence of active construal. In particular, heuristic search over action sequences is another mechanism for reducing the cost of planning, but it accomplishes this in a different way: by examining possible action sequences in order of how promising they seem, not by simplifying the task representation. If people are simulating candidate action sequences via heuristic search (and not engaged in an active construal process), differential attention to task elements could have simply been a side-effect of how those simulations unfolded.

Thus, we wanted to derive predictions of differential awareness as a byproduct of search over action sequences. To do so, we considered two general classes of heuristic search algorithms. The first, a variant of Real-Time Dynamic Programming (RTDP)~\supercite{barto1995learning,bonet2003labeled}, is a \emph{trajectory-based} search algorithm that simulates physically realizable trajectories (i.e., sequences of states and actions that could be generated by repeatedly calling a fixed transition function). The algorithm works by first initializing a heuristic value function (e.g., based on domain knowledge). Then, it simulates trajectories that greedily maximize the heuristic value function while also performing Bellman updates at simulated states~\supercite{barto1995learning}. This scheme then leads RTDP to simulate states in order of how promising they are (according to the continuously updated heuristic value function) until the value function converges. Importantly, RTDP can end up visiting a fraction of the total state space, depending on the heuristic. Our implementation was based on the Labeled RTDP algorithm of Bonet \& Geffner~\supercite{bonet2003labeled}, which additionally includes a labeling scheme that marks states where the estimate of the value function has converged, leading to faster overall convergence.

To derive obstacle awareness predictions, we ran RTDP (implemented in \texttt{msdm}; see Code Availability Statement) on each maze and initialized it with a heuristic corresponding to the optimal value function \textit{assuming there are plus-shaped walls but no obstacles}. This models the background knowledge participants have about distances, while also providing a fair comparison to the initial information provided to the value-guided construal implementation. Additionally, if at any point the algorithm had to choose actions based on estimated value, ties were resolved randomly, making the algorithm stochastic. For each maze, we ran 200 simulations of the algorithm to convergence and examined which states were visited by the algorithm over all simulations. We calculated the mean number of times each obstacle was \textit{hit} by the algorithm, where a hit was defined as a visit to a state adjacent to an obstacle such that the obstacle was in between the state and the goal. Because the distribution of hit counts has a long tail, we used the natural log of hit counts $+ 1$ as the obstacle hit scores. The reason why the raw hit counts have a long tail is due to the particular way in which RTDP calculates the value of regions where the heuristic value is much higher than the actual value (e.g., dead ends in a maze). Specifically, RTDP explores such regions until it has confirmed that it is no better than an alternative path, which can take many steps. More generally, trajectory-based algorithms are limited in that they can only update states by simulating physically realizable trajectories starting from the initial state. 

The limitations of trajectory-based planning algorithms motivated our use of a second class of \textit{graph-based} planning algorithms. We used LAO$^*$~\supercite{hansen2001lao}, a version of the classic A$^*$ algorithm~\supercite{hart1968formal} generalized to be used on Markov Decision Processes (implemented in \texttt{msdm}; see Code Availability Statement). Unlike trajectory-based algorithms, graph-based algorithms like LAO$^*$ maintain a graph of previously simulated states. LAO$^*$ in particular builds a graph of the task rooted at the initial state and then continuously plans over the graph. If it computes a plan that leads it to a state at the edge of the graph, the graph is expanded according to the transition model to include that state and then the planning cycle is restarted. Otherwise, if it computes an optimal plan that only visits states in the simulated graph, the algorithm terminates. By continuously expanding the task graph and performing planning updates, the algorithm can intelligently explore the most promising (according to the heuristic) regions of the state space being constrained to physically realizable sequences. In particular, graph-based algorithms can quickly ``backtrack'' when they encounter dead ends.

Obstacle awareness predictions based on LAO$^*$ were derived by using the same initial heuristic as was used for RTDP and a similar scheme for handling ties. We then calculated the total number of times an obstacle was hit during graph expansion phases only, using the same definition of a hit as above. For each maze, we generated 200 planning simulations and used the raw hit counts as the hit score. 

Algorithms like RTDP and LAO$^*$ plan by simulating realizable action sequences that begin at the start state. As a result, these models tend to predict greater awareness to obstacles that are near the start state and are consistent with the initial heuristic, regardless of whether those obstacles strongly affect or lie along the final optimal path. For instance, obstacles down initially promising dead ends have a high hit score. This contrasts with value-guided construal, which predicts greater attention to relevant obstacles, even if they are distant, and lower attention to irrelevant ones, even if they are nearby. For an example of these distinct model predictions, see maze \#14 in Extended Data Figure 6. 

To be clear, our goal was to obtain predictions for search over action sequences \textit{in the absence of an active construal process} for comparison with value-guided construal. However, in general, heuristic search and value-guided construal are complementary mechanisms, since the former is a way to plan given a representation and the latter is a way to choose a representation for planning. For instance, one could perform heuristic search over a construed planning model, or a construal could help with selecting a heuristic to guide search over actions. These kinds of interactions between action-sequence search and construal are important directions for future research that can be built on the ideas developed here.

\subsubsection*{Successor Representation-based Predictors}
We also considered two measures based on the \textit{successor representation}, which has been proposed as a component in several computational theories of efficient sequential decision-making~\supercite{momennejad2017successor,gershman2018successor}. Importantly, the successor representation is not a specific model; rather it is a predictive coding of a task in which states are represented in terms of the \textit{future states} likely to be visited from that state, given the decision-maker follows a certain policy. Formally, the value function of a policy $\pi(a \mid s)$ can be expressed in the following two equivalent ways:
\begin{align}
V_{\pi}(s) &= U(s) + \sum_a \pi(a \mid s) \sum_{s'}P(s' \mid s, a)V_{\pi}(s') \\
          &= \sum_{s^{+}}M_{\pi}(s, s^{+}) U(s^{+}),
\end{align}
where $M_{\pi}(s, s^{+})$ is expected occupancy of $s^{+}$ starting from $s$, when acting according to $\pi$. The successor representation of a state $s$ under $\pi$ is then the vector $M_{\pi}(s, \cdot)$. Algorithmically, $M_\pi$ can be calculated by solving a set of recursive equations (implemented in Python with \texttt{numpy}~\supercite{numpy2020}; see Code Availability Statement):
\begin{equation}
    M_\pi(s, s^{+}) = 
        \mathbbm{1}[s = s^{+}] + \sum_{a, s'}\pi(a \mid s)P(s' \mid s, a)M_\pi(s',s^{+}).
\end{equation}
Again, the successor representation is not itself an algorithm, but rather a policy-conditioned re-coding of states that can be a component of a larger computational process (e.g, different kinds of learning or planning). Here, we focus on its use in the context of \textit{transfer learning}~\supercite{momennejad2017successor,russek2017predictive} and \textit{bottleneck states}~\supercite{solway2014,stachenfeld2017hippocampus}.

Research on transfer learning posits that the successor representation supports transfer that is more flexible than pure model-free mechanisms but less flexible than model-based planning. For example, Russek et al. \supercite{russek2017predictive} model agents that learned a successor representation for the optimal policy in an initial maze and then examined transfer when the maze was changed (e.g., adding in a new barrier). While their work focuses on learning, rather than planning, we can borrow the basic insight that the successor representation induced by the optimal policy for a source task can influence the encoding of a target task, which constitutes a form of construal. In our experiments, participants were not trained on any particular source task, but we can use the maze with all obstacles removed as a proxy (i.e., representing what all mazes had in common). Thus, we calculated the optimal policy $\pi$ for the maze without any obstacles (but with the start and goal), computed the successor representation $M_\pi$, and then calculated, for each obstacle $i$ in the actual maze with the obstacles, a \textit{successor representation overlap} (SR-Overlap) score:
\begin{equation}
    \text{SR-Overlap}(i) = \sum_{s \in \text{Obs}_i} M_{\pi}(s_0, s),
\end{equation}
where $s_0$ is the starting state and Obs$_i$ is the set of states occupied by the obstacle $i$. This quantity can be interpreted as the amount of overlap between an obstacle and the successor representation of the starting state. If the successor representation shapes how people represent tasks, this quantity would be associated with greater awareness of certain obstacles.

The second predictor is related to the idea of \textit{bottleneck states}. These emerge from how the successor representation encodes multi-scale task structure~\supercite{stachenfeld2017hippocampus}, and they have been proposed as a basis for subgoal selection~\supercite{solway2014}. If bottlenecks guide subgoal selection, then distance to bottleneck states could give rise to differential awareness of obstacles via subgoaling processes. Thus, we wanted to test that responses consistent with value-guided construal were not entirely attributable to the effect of bottleneck states calculated in the absence of an active construal process. Importantly, we note that as with alternative planning mechanisms like heuristic search, the identification of bottleneck states for subgoaling is compatible with value-guided construal (e.g., one could identify subgoals for a construed version of a task).

When viewing the transition function of a task (e.g., a maze) as a graph over states, bottleneck states lie on either side of a partitioning of the state space into two regions such that there is high intra-region connectivity and low inter-region connectivity. This can be computed for any transition function using the normalized min-cuts algorithm~\supercite{shi2000normalized} or derived from the second eigenvector of the successor representation under a random policy~\supercite{stachenfeld2017hippocampus}. Here, we use a variant of the second approach as described in the appendix of~\supercite{stachenfeld2017hippocampus}. Formally, given a transition function, $P(s' \mid s, a)$, we define an adjacency matrix, $A(s, s') = \mathbbm{1}[\exists a \text{ s.t. } P(s' \mid s, a) > 0]$, and a diagonal degree matrix, $D(s, s) = \sum_{s'}A(s, s')$. Then, the graph Laplacian, a representation often used to derive low-dimensional embeddings of graphs in spectral graph theory, is $L = D - A$. We take the eigenvector with the second largest eigenvalue, which assigns a positive or negative value to each state in the task. This vector can be interpreted as projecting the state space onto a single dimension in a way that best preserves connectivity information, with a zero point that represents the mid-point of the projected graph. Bottleneck states correspond to those states nearest to 0. For each maze, we used this method to identify bottleneck states and further reduced these to the \textit{optimal bottleneck states}, defined as bottleneck states with a non-zero probability of being visited under the optimal stochastic policy for the maze. Finally, for each obstacle, we calculated a \textit{bottleneck distance} score, the minimum Manhattan distance from an obstacle to any of these bottleneck states.

Notably, value-guided construal also predicts greater attention to obstacles that form bottlenecks because one often needs to carefully navigate through them to reach the goal. However, the predictions of our model differ for obstacles that are \textit{distant} from the bottleneck. Specifically, value-guided construal predicts greater attention to relevant obstacles that affect the optimal plan, even if they are far from the bottleneck (e.g., see model predictions for maze \#2 in Extended Data Figure 5).

\subsubsection*{Perceptual Landmarks}
Finally, we considered several predictors based on low-level perceptual landmarks and participants' behavior. These included the minimum Manhattan distance from an obstacle to the start location, the goal location, the center black walls, the center of the grid, and any of the locations visited by the participant in a trial (navigation distance).  We also considered the timestep at which participants were closest to an object as a measure of how recently they were near an object. In cases where navigation distance was not an appropriate measure (e.g., if participants never navigated to the goal), we used the minimum Manhattan distance to trajectories sampled from the optimal policy averaged over 100 samples.

\subsection*{Experimental Design}

All experiments were pre-registered (see Data Availability Statement) and approved by the Princeton Institutional Review Board (IRB). All participants were recruited from the Prolific online platform and provided informed consent. At the end of each experiment, participants provided free-response demographic information (age and gender, coded as male/female/neither). Experiments were implemented with psiTurk~\supercite{gureckis2016psiturk} and jsPsych~\supercite{de2015jspsych} frameworks (see Code Availability Statement). Instructions and example trials are shown in the Supplementary Experimental Materials. 

\subsubsection*{Initial experiment}
Our initial experiment used a maze-navigation task in which participants moved a circle from a starting location on a grid to a goal location using the arrow keys. The set of \textit{initial mazes} consisted of twelve 11 $\times$ 11 mazes with seven blue tetronimo-shaped obstacles and center walls arranged in a cross that blocked movement. On each trial, participants were first shown a screen displaying only the center walls. When they pressed the spacebar, the circle they controlled, the goal, and the obstacles appeared, and they could begin moving immediately. In addition, to ensure that participants remained focused on moving, we placed a green square on the goal that shrank and would disappear after 1000ms but reset whenever an arrow key was pressed, except at the beginning of the trial when the green square took longer to shrink (5000ms). Participants received \$0.10 for reaching the goal without the green square disappearing (in addition to the base pay of \$0.98). The mazes were pseudo-randomly rotated or flipped, so the start and end state was constantly changing, and the order of mazes were pseudo-randomized.  After completing each trial, participants received awareness probes, which showed a static image of the maze they had just navigated, with one of the obstacles shown in light blue. They were asked ``How aware of the highlighted obstacle were you at any point?'' and could respond using an 8-point scale (rescaled from 0 to 1 for analyses). Probes were presented for the seven obstacles in a maze. None of the probes were associated with a bonus.

We requested 200 participants on Prolific and received 194 complete submissions. Following pre-registered exclusion criteria, a trial was excluded if, during navigation, $>$5000ms was spent at the initial state, $>$2000ms was spent at any non-initial state, $>$20000ms was spent on the entire trial, or $>$1500ms was spent in the last three steps in total. Participants with $<$80\% of trials after exclusions or who failed 2 of 3 comprehension questions were excluded, which resulted in $n=161$ participants' data being analyzed (median age of $28$; $81$ male, $75$ female, $5$ neither). 

\subsubsection*{Up-front planning experiment}
The \textit{up-front planning} version of the memory experiment was designed to dissociate planning and execution. The main change was that after participants took their first step, all of the blue obstacles (but not the walls or goal) were removed from the display (though they still blocked movement). This strongly encouraged planning prior to execution. To provide sufficient time to plan, the green square took 60000ms to shrink on the first step. Additionally, on a random half of the trials, after taking two steps, participants were immediately presented with the awareness probes (\textit{early termination} trials). The other half were \textit{full} trials. We reasoned that responses following early termination trials would better reflect awareness after planning but before execution (see Supplementary Memory Experiment Analyses for analyses comparing early versus full trials). 

We requested 200 participants on Prolific and received 188 complete submissions. The exclusion criteria were the same as in the initial experiment, except that the initial state and total trial time criteria were raised to 30000ms and 60000ms, respectively. After exclusions, we analyzed data from $n=162$ participants (median age of 28; 85 male, 72 female, 5 neither).

\subsubsection*{Critical mazes experiment}
In the \textit{critical mazes experiment}, participants again could not see the obstacles while executing and so needed to plan up front, but no trials ended early. There were two main differences with the previous experiments. First, we used a set of four \textit{critical mazes} that included \textit{critical obstacles} chosen to test predictions specific to value-guided construal. These were obstacles relevant to decision-making, but distant from the optimal path (see Supplementary Memory Experiment Analyses for analyses focusing on these critical obstacles). Second, half of the participants received \textit{recall probes} in which they were shown a static image of the grid with only the walls, a green obstacle, and a yellow obstacle. They were then asked ``An obstacle was either in the yellow or green location (not both), which one was it?'' and could select either option, followed by a confidence judgment on an 8-point scale (rescaled from 0 to 1 for analyses). Pairs of obstacles and their contrasts in the critical mazes are shown in {Extended Data Figure 4a}. Participants each received two blocks of the four critical mazes, pseudo-randomly oriented and/or flipped. 

We requested 200 participants on Prolific and received 199 complete submissions. The trial and participant exclusion criteria were the same as in the up-front planning experiment. After exclusions, we analyzed data from $n=156$ participants (median age of $26$; 78 male, 75 female, 3 neither).

\subsubsection*{Control Experiments}
The aim of the control experiments was to obtain yoked baselines for perception and execution for comparison with probe responses in the memory studies. The \textit{perceptual control} used a variant of the task in which participants were shown patterns that were perceptually identical to the mazes. Instead of solving a maze, they were told to ``catch the red dot'': On each trial, a small red dot could appear anywhere on the grid, and participants were rewarded based on whether they pressed the spacebar after it appeared. Each participant was yoked to the responses of a participant from either the \textit{up-front planning} or \textit{critical mazes} experiments. On \textit{yoked trials}, participants were shown the exact same maze/pattern as their counterpart. Additionally, they were shown the pattern for the amount of time that their counterpart took before making their first move---since the obstacles were not visible during execution for the counterpart, this is roughly the time the counterpart spent looking at the maze to plan. A red dot never appeared on these trials, and they were followed by the exact same probes that the counterpart received. References to ``obstacles'' were changed to ``tiles'' (e.g., ``highlighted tiles'' as opposed to ``highlighted obstacle'' for the awareness probes). We also included \textit{dummy trials}, which showed mazes in orientations not appearing in the yoked trials, for durations sampled from the yoked durations. Half of the dummy trials had red dots. We recruited enough participants such that at least one participant was matched to each participant from the original experiments and excluded people who said that they had participated in a similar experiment. This resulted in data from $n=164$ participants being analyzed for the initial mazes perceptual control (median age of $30.5$; 84 male, 79 female, 1 neither) and $n=172$ for the critical mazes perceptual control (median age of 36.5; 86 male, 85 female, 1 neither).

The \textit{execution control} used a variant of the task in which participants followed a series of ``breadcrumbs'' through the maze to the goal and so did not need to plan a path to the goal. Each participant was yoked to a counterpart in either the initial experiment or the critical mazes experiment so that the breadcrumbs were generated based on the exact path taken by the counterpart. The ordering of the mazes and obstacle probes (i.e., awareness or location recall) were also the same. We recruited participants until at least one participant was matched to each participant from the original experiments. Additionally, we used the same exclusion criteria as in the initial experiment with the additional requirement that all black dots be collected on a trial. This resulted in data from $n=163$ participants being analyzed for the initial mazes execution control (median age of 29; 86 male, 77 female) and $n=161$ for the critical mazes execution control (median age of 30; 94 male, 63 female; 4 neither).  

\subsubsection*{Process-Tracing Experiments}
We ran process-tracing experiments using the initial mazes and the critical mazes. These experiments were similar to the memory experiments, except they used a novel process-tracing paradigm designed to externalize the planning process. Specifically, participants never saw all the obstacles in the maze at once. Rather, at the beginning of a trial, after clicking on a red X in the center of the maze, the goal and agent appeared, and participants could use their mouse to hover over the maze and reveal individual obstacles. An obstacle would become completely visible if the mouse hovered over any tile that was part of it for at least 25ms, until the mouse was moved to a tile that was not part of that obstacle. Once the participant started to move using the arrow keys, the cursor became temporarily invisible (to prevent using the cursor as a cue to guide execution), and the obstacles could no longer be revealed. We examined two dependent measures for each obstacle: whether participants hovered over an obstacle, and if so, the duration of hovering in log milliseconds.

For each experiment with each set of mazes, we requested 200 participants on Prolific. Participants who completed the task had their data excluded if they did not hover over any obstacles on more than half of the trials. For the experiment with the initial set, we received completed submissions from 174 people and, after exclusions, analyzed data from $n = 167$ participants (median age of 30; 82 male, 82 female, 3 neither). For the experiment with the critical set, we received completed submissions from 188 people and, after exclusions, analyzed data from $n=179$ participants (median age of 32; 89 male, 86 female, 4 neither).

\subsection*{Experiment Analyses}
Hierarchical generalized linear models (HGLMs) were implemented in Python and R using the \texttt{lme4}~\supercite{lme42015} and \texttt{rpy2}~\supercite{rpy2_2020} packages (see Code Availability Statement). For all models, we included by-participant and by-maze random intercepts, unless the resulting model was singular, in which case we removed by-maze random intercepts. For the memory experiment analyses testing whether value-guided construal predicted responses, we fit models with and without z-score normalized value-guided construal probabilities as a fixed effect and performed likelihood ratio tests to assess significance. For the control experiment analyses reported in the main text, we calculated mean by-obstacle responses from the perceptual and execution controls, and then included these values as fixed effects in models fit to the responses in the planning experiments. We then contrasted models with and without value-guided construal and performed likelihood ratio tests (additional analyses are reported in the Supplementary Memory Experiment Analyses and Supplementary Control Experiment Analyses).

For our comparison with alternative models, we considered 11 different predictors that assign scores to obstacles in each maze: fixed-parameter value-guided construal modification probability (VGC), trajectory-based heuristic search score (Traj HS), graph-based heuristic search score (Graph HS), bottleneck state distance (Bottleneck), successor representation overlap (SR Overlap), minimum navigation distance (Nav Dist), timestep of minimum navigation distance (Nav Dist Step), minimum optimal policy distance (Opt Dist), distance to goal (Goal Dist), distance to start (Start Dist), distance to center walls (Wall Dist), and distance to the center of the maze (Center Dist). We included predictors in the analysis of each experiment's data where appropriate. For example, in the \textit{up-front planning} experiment, participants did not navigate on early termination trials, and so we used the optimal policy distance rather than navigation distance. All predictors were z-score normalized before being included as fixed effects in HGLMs in order to facilitate comparison of estimated coefficients.

We performed three types of analyses using the 11 predictors. First, we wanted determine whether value-guided construal captured variability in responses from the planning experiments even when accounting for the other predictors. For these analyses, we compared HGLMs that included all predictors to HGLMs with all predictors except value-guided construal and tested whether there was a significant difference in fit using likelihood ratio tests (Extended Data Table 1). Second, we wanted to evaluate the relative necessity of each mechanism for explaining attention to obstacles when planning. For these analyses, we compared global HGLMs to HGLMs with each of the predictors removed and calculated the resulting change in AIC (see Extended Data Table 1 for estimated coefficients and resulting AIC values). Finally, we wanted to assess the relative sufficiency of predictors in accounting for responses on the planning tasks. For these analyses, we fit HGLMs to each set of responses that included only individual predictors or pairs of predictors, and for each model we calculated the $\Delta$AIC relative to the best-fitting model (Extended Data Figure 8). Note that for all of these models, AIC values are summed over participants. 

\printbibliography[title={Methods References},segment=\therefsegment,notcategory=inMain]

\newpage

\begin{justify}
\noindent {\bf Acknowledgements}:
The authors would like to thank Jessica Hamrick, Louis Gularte, Ceyda Sayal{\i}, Qiong Zhang, Rachit Dubey, and William Thompson for valuable feedback on this work. This work was funded by NSF grant \#1545126, John Templeton Foundation grant \#61454, and AFOSR grant \# FA 9550-18-1-0077. 
\newline

\noindent {\bf Author Contributions}:
All authors contributed to conceptualizing the project and editing the manuscript. MKH, DA, MLL, and TLG developed the value-guided construal model. MKH implemented it. MKH and CGC implemented the heuristic search models and \texttt{msdm} library. MKH, JDC, and TLG designed the experiments. MKH implemented the experiments, analyzed the results, and drafted the manuscript.
\newline

\noindent {\bf Competing Interest Declaration}:
The authors declare no competing interests.
\newline

\noindent {\bf Supplementary Information} is available for this paper.
\newline

\noindent {\bf Data Availability Statement}: Data for the current study are available through the Open Science Foundation repository \href{http://doi.org/10.17605/OSF.IO/ZPQ69}{http://doi.org/10.17605/OSF.IO/ZPQ69}.
\newline

\noindent {\bf Code Availability Statement}: Code for the current study are available through the Open Science Foundation repository \href{http://doi.org/10.17605/OSF.IO/ZPQ69}{http://doi.org/10.17605/OSF.IO/ZPQ69}, which links to a GitHub repository and contains an archived version of the repository. The value-guided construal model and alternative models were implemented in Python (3.7) using the \texttt{msdm} (0.6) library, \texttt{numpy} (1.19.2), and \texttt{scipy} (1.5.2). Experiments were implemented using \texttt{psiTurk} (3.2.0) and \texttt{jsPsych} (6.0.1). Hierarchical generalized linear regressions were implemented using \texttt{rpy2} (3.3.6), \texttt{lme4} (1.1.21), and R (3.6.1).
\end{justify}

\newpage

\begin{figure}[H]
    \centering
    \includegraphics[height=.9\textheight]{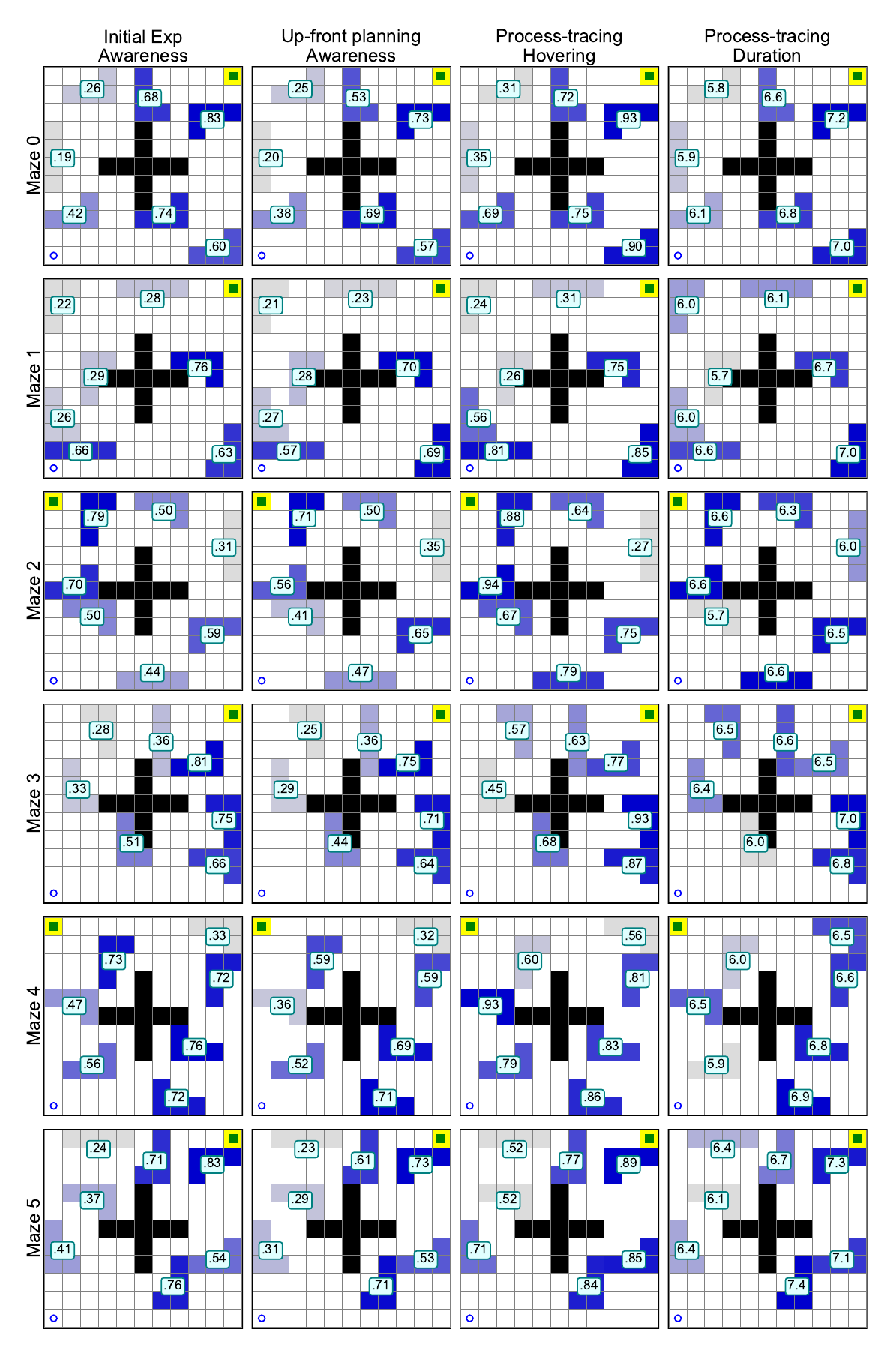}
\end{figure}
\textbf{Extended Data Fig. 1 $\mid$ Experimental measures on mazes 0 to 5,} Average responses associated with each obstacle in mazes 0 to 5 in the initial experiment (awareness judgment), the up-front planning experiment (awareness judgment), and the process-tracing experiment (whether an obstacle was hovered over and, if so, the duration of hovering in log milliseconds). Obstacle colors are normalized by the minimum and maximum values for each measure/maze, except for awareness judgments, which are scaled from 0 to 1.

\newpage
\begin{figure}[H]
    \centering
    \includegraphics[height=.9\textheight]{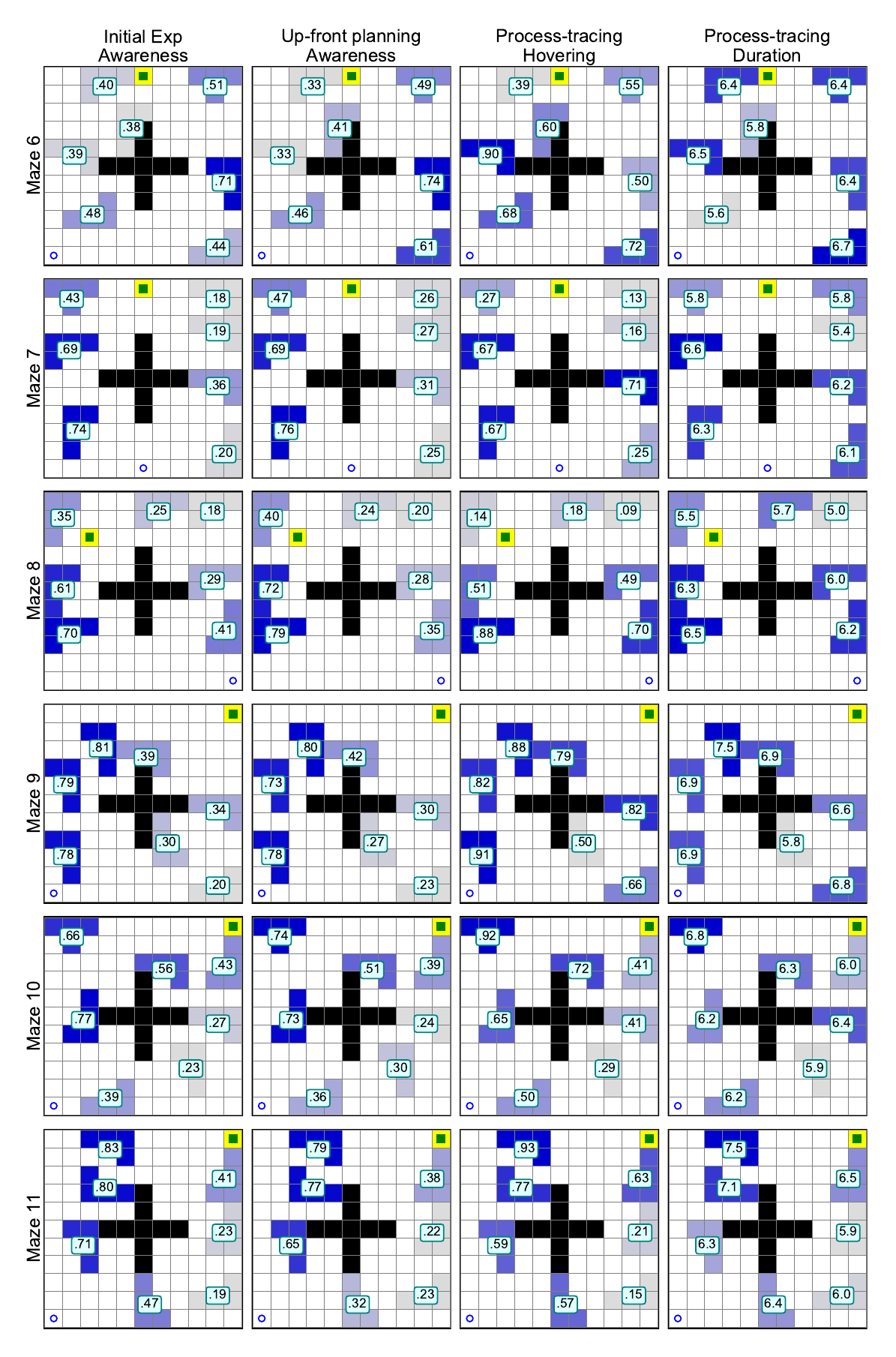}
\end{figure}
\textbf{Extended Data Fig. 2 $\mid$ Experimental measures on mazes 6 to 11,} Average responses associated with each obstacle in mazes 6 to 11 in the initial experiment (awareness judgment), the up-front planning experiment (awareness judgment), and the process-tracing experiment (whether an obstacle was hovered over and, if so, the duration of hovering in log milliseconds). Obstacle colors are normalized by the minimum and maximum values for each measure/maze, except for awareness judgments, which are scaled from 0 to 1.

\newpage
\begin{figure}[H]
    \centering
    \includegraphics[width=\textwidth]{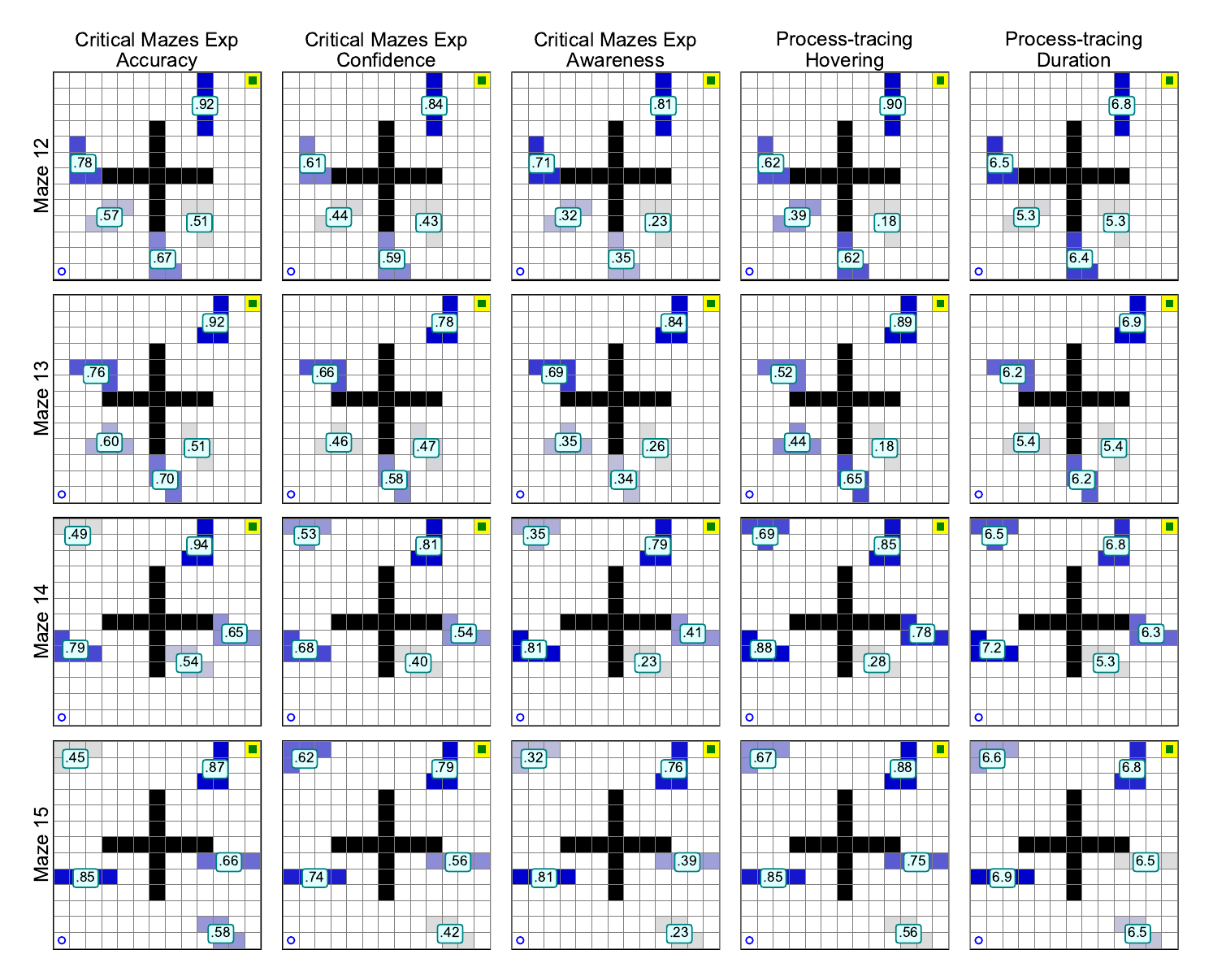}
\end{figure}
\textbf{Extended Data Fig. 3 $\mid$ Experimental measures on mazes 12 to 15,} Average responses associated with each obstacle in mazes 12 to 15 in the critical mazes experiment (recall accuracy, recall confidence, and awareness judgment) and the process-tracing experiment (whether an obstacle was hovered over and, if so, the duration of hovering in log milliseconds). Obstacle colors are scaled to range from 0.5 to 1.0 for accuracy, 0 to 1 for hovering, confidence, and awareness judgments, and the minimum to maximum values across obstacles in a maze for hovering duration in log milliseconds. 

\newpage
\begin{figure}[H]
    \centering
    \includegraphics[width=\textwidth]{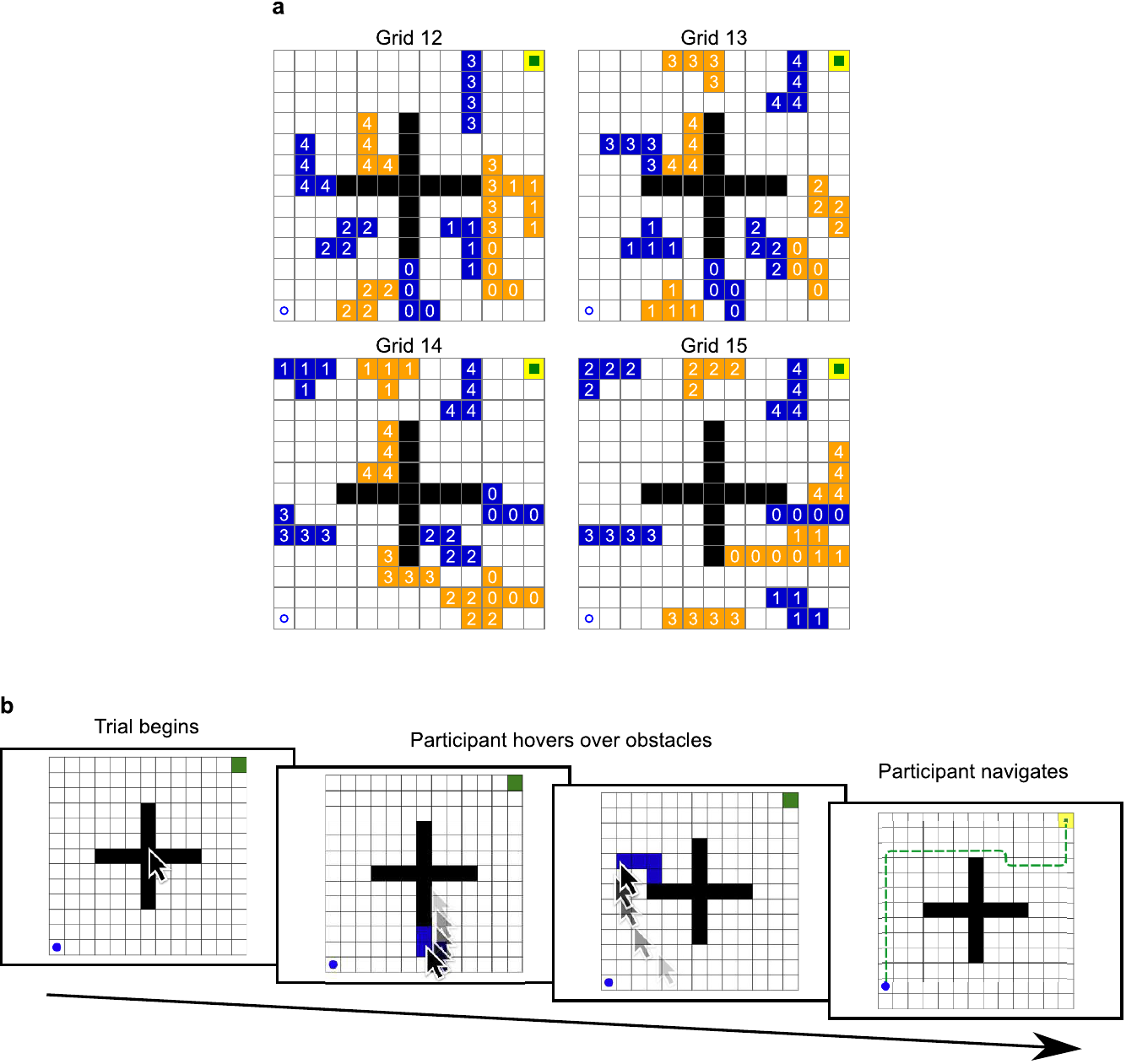}
\end{figure}
\textbf{Extended Data Fig. 4 $\mid$ Additional Experimental Details,} \textbf{a,} Items from critical mazes experiment. Blue obstacles are the location of obstacles during the navigation part of the trial. Orange obstacles with corresponding number are copies that were shown during location recall probes. During recall probes, participants only saw an obstacle paired with its copy. \textbf{b,} Example trial from process-tracing experiment. Participants could never see all the obstacles at once, but, before navigating, could use their mouse to reveal obstacles. We analyzed whether value-guided construal predicted which obstacles people tended to hover over and, if so, the duration of hovering.

\newpage
\begin{figure}[H]
    \centering
    \includegraphics[height=.9\textheight]{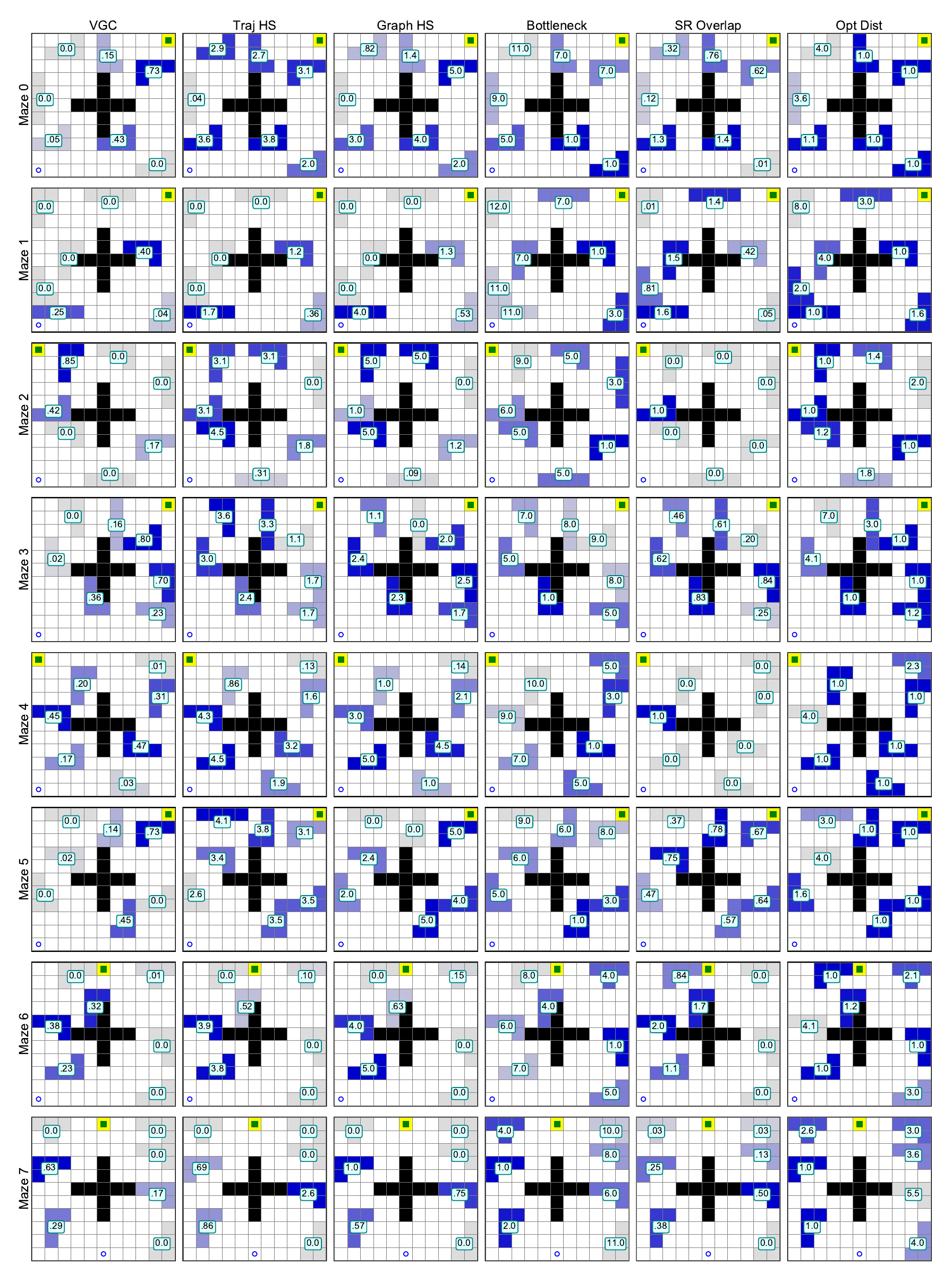}
\end{figure}
\textbf{Extended Data Fig. 5 $\mid$ Model predictions on mazes 0 through 7,} Shown are the predictions for six of the eleven predictors we tested: fixed parameter value-guided construal modification obstacle probability (VGC, our model); trajectory-based heuristic search obstacle hit score (Traj HS); graph-based heuristic search obstacle hit score (Graph HS); distance to optimal bottleneck (Bottleneck); successor representation overlap score (SR Overlap); and distance to optimal paths (Opt Dist) (see Methods, Model Implementations). Mazes 0 to 7 were all in the initial set of mazes. Darker obstacles correspond to greater predicted attention according to the model. Obstacle colors normalized by the minimum and maximum values for each model/maze.

\newpage
\begin{figure}[H]
    \centering
    \includegraphics[height=.9\textheight]{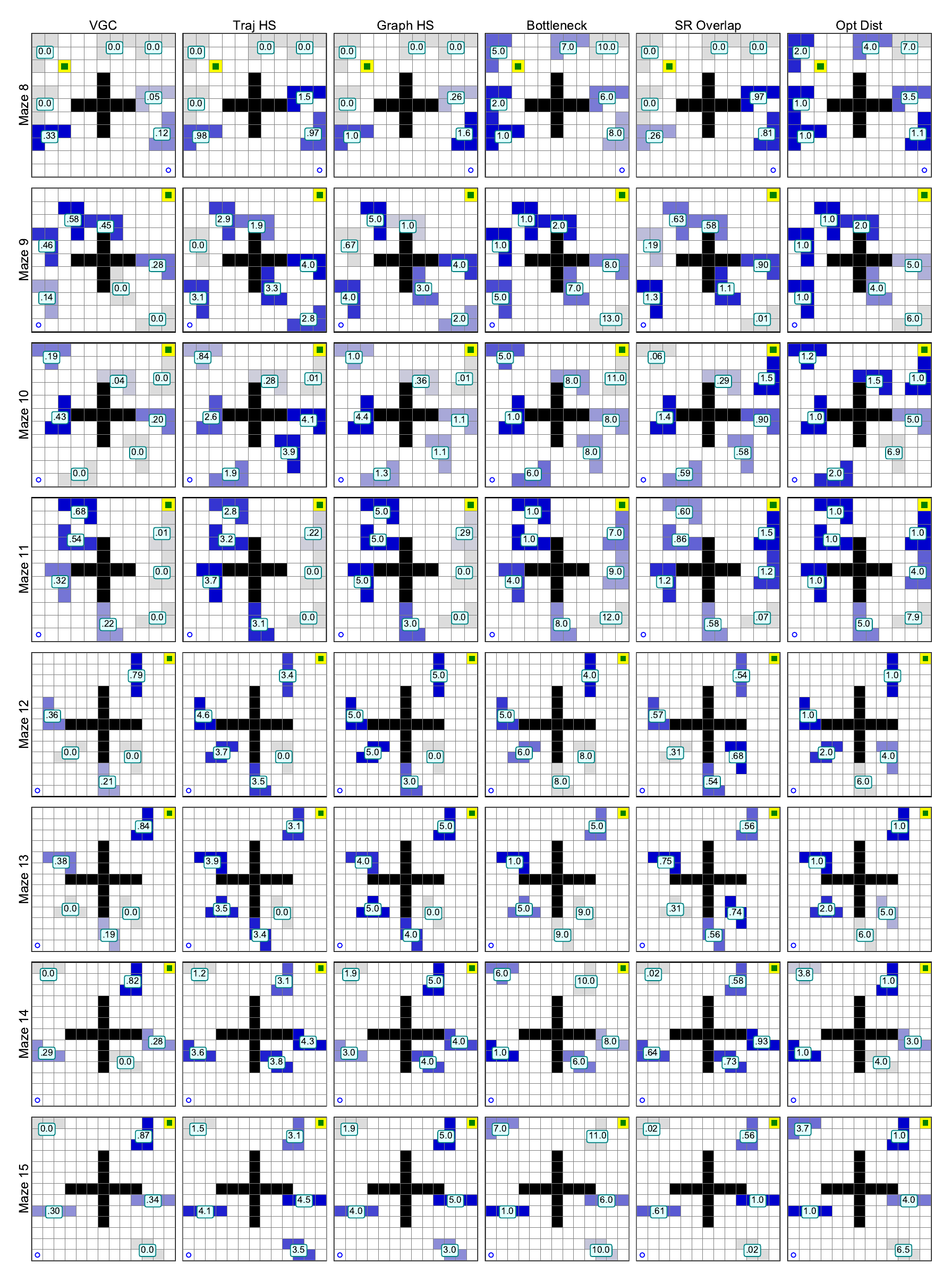}
\end{figure}
\textbf{Extended Data Fig. 6 $\mid$ Model predictions on mazes 8 through 15,} Shown are the predictions for six of the eleven predictors we tested (see Methods, Model Implementations). Mazes 8 to 11 were part of the initial set of mazes, while mazes 12 to 15 constituted the set of critical mazes. Darker obstacles correspond to greater predicted attention according to the model. Obstacle colors normalized by the minimum and maximum values for each model/maze.

\newpage
\begin{figure}[H]
    \centering
    \includegraphics[width=\textwidth]{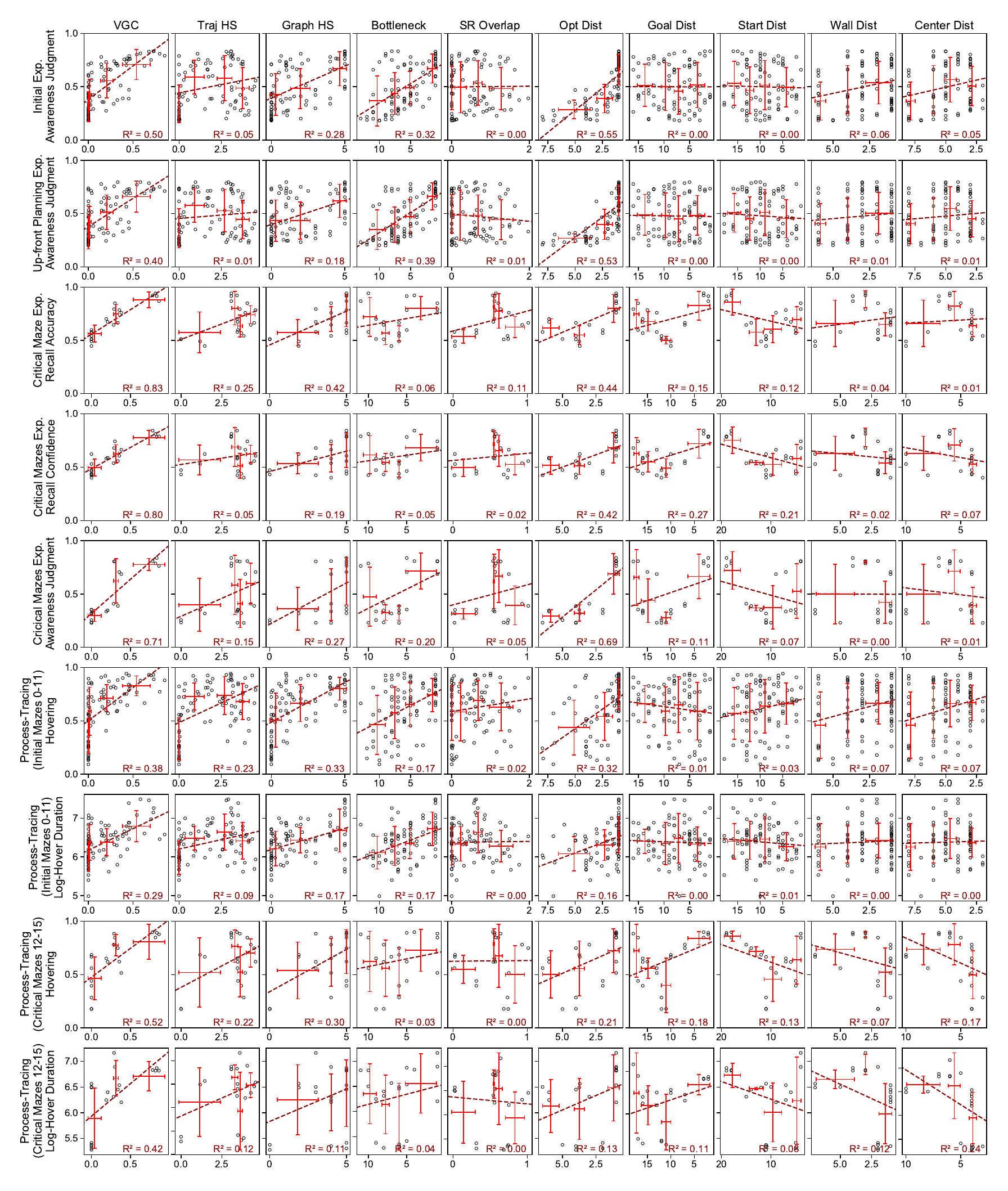}
\end{figure}
\vspace{-2em}
\textbf{Extended Data Fig. 7 $\mid$ Summaries of candidate models and data from planning experiments,} Each row corresponds to a measurement of attention to obstacles from a planning experiment: Awareness judgments from the initial memory experiment, the up-front planning experiment, and the critical mazes experiment; recall accuracy and confidence from the critical mazes experiment; and the binary hovering measure and hovering duration measure (in log milliseconds) from the two process-tracing experiments. Each column corresponds to candidate processes that could predict attention to obstacles: fixed parameter value-guided construal modification obstacle probability (VGC, our model), trajectory-based heuristic search hit score (Traj HS), graph-based heuristic search hit score (Graph HS), distance to bottleneck states (Bottleneck), successor-representation overlap (SR Overlap), expected distance to optimal paths (Opt Dist), distance to the goal location (Goal Dist), distance to the start location (Start Dist), distance to the invariant black walls (Wall Dist), and distance to the center of the maze (Center Dist). Note that for distance-based predictors, the x-axis is flipped. For each predictor, we quartile-binned the predictions across obstacles, and for each bin we plot (bright red lines) the mean and standard deviation of the predictor and mean by-obstacle response (overlapping bins were collapsed into a single bin). Black circles correspond to the mean response and prediction for each obstacle in each maze. Dashed dark red lines are simple linear regressions on the black circles, with $R^2$ values shown in the lower right of each plot. Across the nine measures, value-guided construal tracks attention to obstacles, while other candidate processes are less consistently associated with obstacle attention (data are based on $n = 84215$ observations taken from $825$ independent participants).

\begin{figure}[H]
    \centering
    \includegraphics[width=\textwidth]{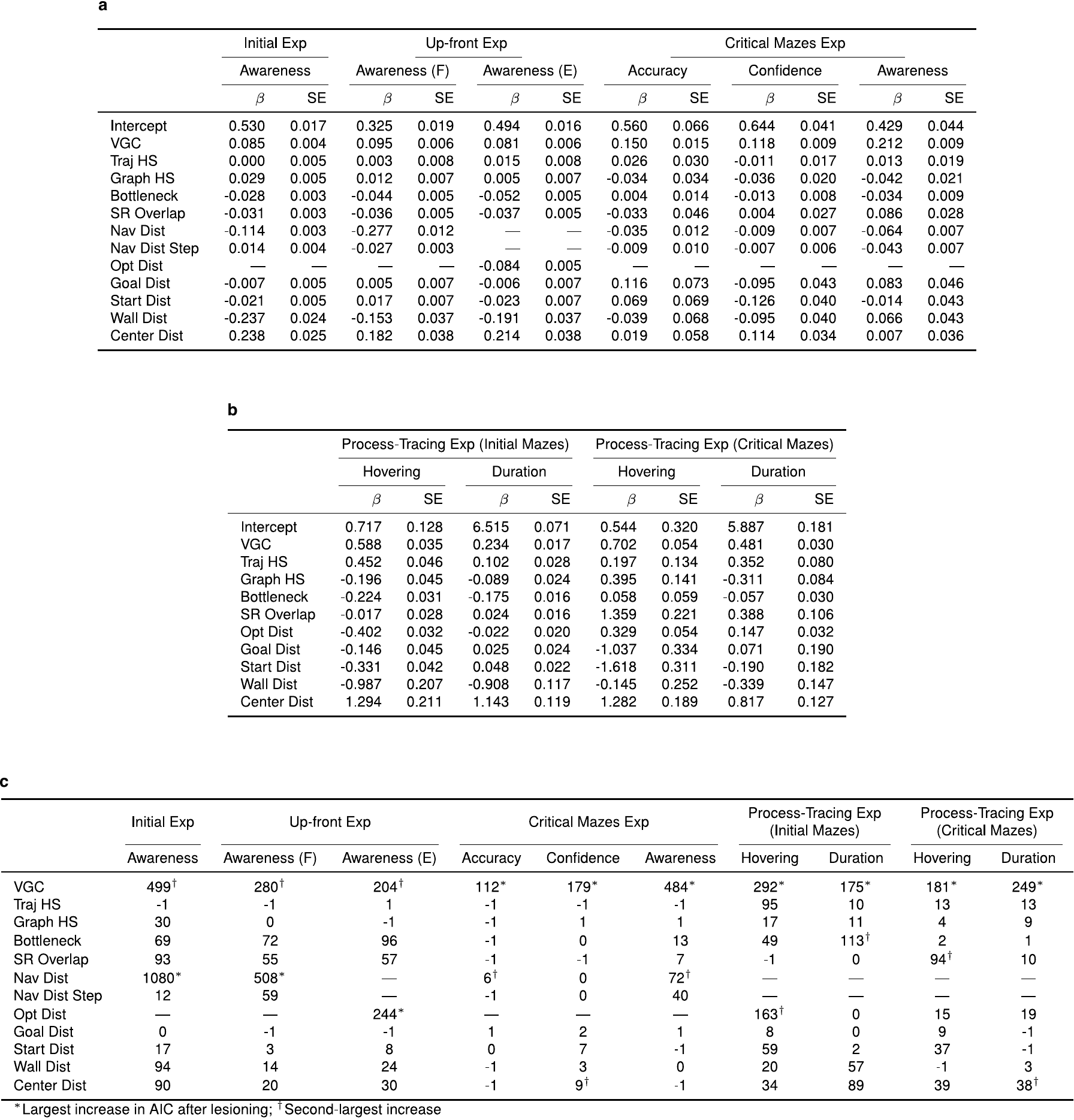}
\end{figure}
\textbf{Extended Data Table 1 $\mid$ Necessity of different mechanisms for explaining attention to obstacles when planning,} For each measure in each planning experiment, we fit hierarchical generalized linear models (HGLMs) that included the following predictors as fixed-effects: fixed parameter value-guided construal modification obstacle probability (VGC, our model); trajectory-based heuristic search obstacle hit score (Traj HS); graph-based heuristic search obstacle hit score (Graph HS); distance to optimal bottleneck (Bottleneck); successor representation overlap score (SR Overlap); distance to path taken (Nav Dist); timestep of point closest along path taken (Nav Dist Step); distance to optimal paths (Opt Dist); distance to the goal state (Goal Dist); distance to the start state (Start Dist); distance to any part of the center walls (Wall Dist); and distance to the center of the maze (Center Dist) (Methods, Model Implementations). If the measure was taken before participants navigated, distance to the optimal paths was used, otherwise, distance to the path taken and its timestep were used. \textbf{a, b, } Estimated coefficients and standard errors for z-score normalized predictors in HGLMs fit to responses from the initial experiment, up-front planning experiment (F = full trials, E = early termination trials), the critical mazes experiment, and the process-tracing experiments. We found that value-guided construal was a significant predictor even when accounting for alternatives (likelihood ratio tests between full global models and models without value-guided construal: Initial Exp, Awareness: $\chi^2(1) = 501.11, p < 1.0 \times 10^{-16}$; 
Up-front Exp, Awareness (F): $\chi^2(1) = 282.17, p < 1.0 \times 10^{-16}$; 
Up-front Exp, Awareness (E): $\chi^2(1) = 206.14, p < 1.0 \times 10^{-16}$; 
Critical Mazes Exp, Accuracy: $\chi^2(1) = 114.87, p < 1.0 \times 10^{-16}$; 
Critical Mazes Exp, Confidence: $\chi^2(1) = 181.28, p < 1.0 \times 10^{-16}$; 
Critical Mazes Exp, Awareness: $\chi^2(1) = 486.99, p < 1.0 \times 10^{-16}$; 
Process-Tracing Exp (Initial Mazes), Hovering: $\chi^2(1) = 294.40, p < 1.0 \times 10^{-16}$; 
Process-Tracing Exp (Initial Mazes), Duration: $\chi^2(1) = 177.58, p < 1.0 \times 10^{-16}$; 
Process-Tracing Exp (Critical Mazes), Hovering: $\chi^2(1) = 183.52, p < 1.0 \times 10^{-16}$; 
Process-Tracing Exp (Critical Mazes), Duration: $\chi^2(1) = 251.16, p < 1.0 \times 10^{-16}$\unskip). \textbf{c, } To assess the relative necessity of each predictor for the fit of a HGLM, we conducted lesioning analyses in which, for each predictor in a given \textit{global} HGLM, we fit a new \textit{lesioned} HGLM with only that predictor removed. Each entry of the table shows the change in AIC when comparing global and lesioned HGLMs, where larger positive values indicate a greater reduction in fit as a result of removing a predictor. According to this criterion, across all experiments and measures, value-guided construal is either the first or second most important predictor. $^*$Largest increase in AIC after lesioning; $^\dagger$Second-largest increase.

\newpage
\begin{figure}[H]
    \centering
    \includegraphics[height=.92\textheight]{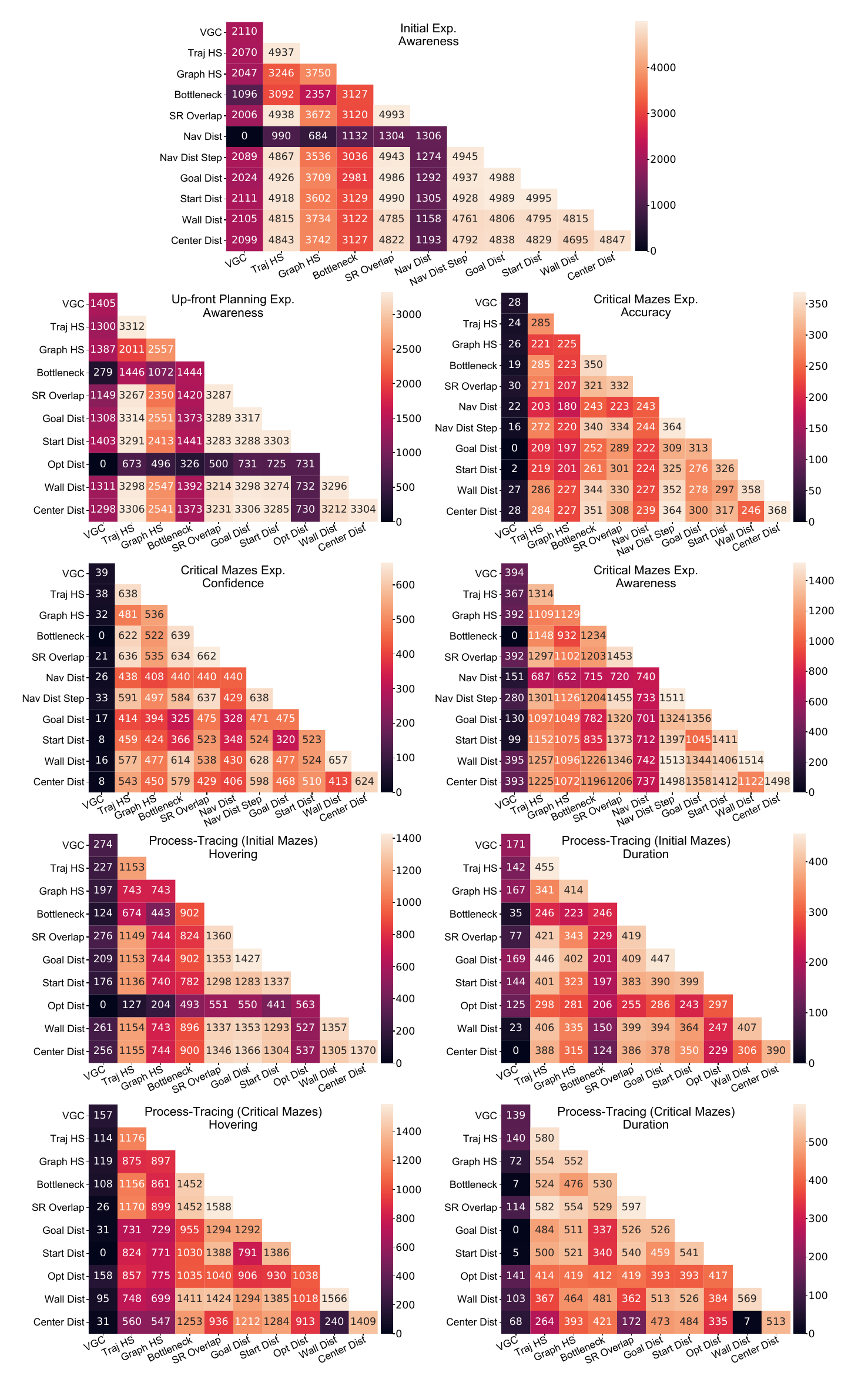}
\end{figure}
\vspace{-2em}
\textbf{Extended Data Figure 8 $\mid$ Sufficiency of individual and pairs of mechanisms for explaining attention to obstacles when planning,} To assess the individual and pairwise sufficiency of each predictor for explaining responses in the planning experiments, we fit hierarchical generalized linear models (HGLMs) that included pairs of predictors as fixed effects. Each lower-triangle plot corresponds to one of the experimental measures, where pairs of predictors included in a HGLM as fixed-effects are indicated on the x- and y-axes. Values are the $\Delta$AIC for each model relative to the best fitting model associated with an experimental measure (lower values indicate better fit). Values along the diagonals correspond to models fit with a single predictor. According to this criterion, across all experimental measures, value-guided construal is the first, second, or third best single-predictor HGLM, and is always in the best two-predictor HGLM.

\newpage
\begin{figure}[H]
    \centering
    \includegraphics[width=.9\textwidth]{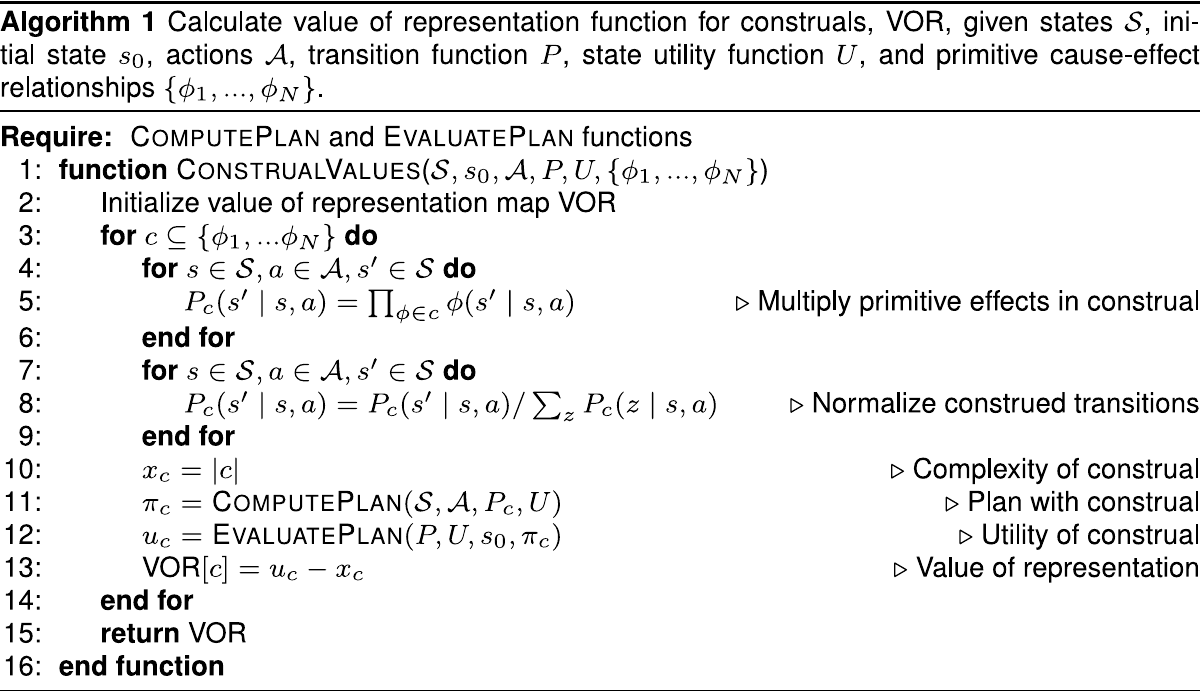}
\end{figure}
\textbf{Extended Data Table 2 $\mid$ Algorithm for Computing the Value of Representation Function} To obtain predictions for our our ideal model of value-guided construal, we calculated the value of representation of all construals in a maze. This was done by enumerating all construals (subsets of obstacle effects) and then, for each construal, calculating its behavioral utility and cognitive cost. This allows us to obtain theoretically optimal value-guided construals. For a discussion of alternative ways of calculating construals, see the Supplementary Discussion of Construal Optimization Algorithms.

\end{document}